\documentclass{article}

\usepackage[final]{neurips_2023}

\usepackage[utf8]{inputenc} % allow utf-8 input
\usepackage[T1]{fontenc}    % use 8-bit T1 fonts
\usepackage{hyperref}       % hyperlinks
\usepackage{url}            % simple URL typesetting
\usepackage{booktabs}       % professional-quality tables
\usepackage{amsfonts}       % blackboard math symbols
\usepackage{nicefrac}       % compact symbols for 1/2, etc.
\usepackage{microtype}      % microtypography
\usepackage{xcolor}         % colors

\usepackage{graphicx}
\graphicspath{{figures/}}
\usepackage{bm}

\usepackage{amsmath}
\usepackage{amssymb}
\usepackage{mathtools}
\usepackage{amsthm}

\usepackage[capitalize,noabbrev]{cleveref}

\theoremstyle{plain}

\theoremstyle{definition}

\theoremstyle{remark}

\usepackage{multirow}

\def\omg{{\Omega}}
\newcommand{\mcU}{\mathcal{U}}

\newcommand{\mcD}{\mathcal{D}}
\newcommand{\mcG}{\mathcal{G}}

\newcommand{\mcL}{\mathcal{L}}

\newcommand{\mcA}{\mathcal{A}}

\def \gb{\bm{g}}
\def \ub{\bm{u}}

\def \vb{\bm{v}}

\def \xb{\bm{x}}

\def \yb{\bm{y}}

\def \hb{\bm{h}}
\def \pb{\bm{p}}
\def \qb{\mathbf{q}}

\def \cb{\bm{c}}

\newcommand{\real}{\mathbb{R}}

\title{Domain Agnostic Fourier Neural Operators}
\author{%
  Ning~Liu\thanks{Equal contribution} \\
  Global Engineering and\\Materials, Inc.\\
  Princeton, NJ 08540 \\
  \texttt{ningliu@umich.edu} \\
  \And
  Siavash~Jafarzadeh$^{*}$ \\
  Department of Mathematics\\
  Lehigh University\\
  Bethlehem, PA 18015 \\
  \texttt{sij222@lehigh.edu} \\
  \And
  Yue~Yu\thanks{Corresponding author} \\
  Department of Mathematics\\
  Lehigh University\\
  Bethlehem, PA 18015 \\
  \texttt{yuy214@lehigh.edu} \\
}

\begin{document}

\maketitle

\begin{abstract}
Fourier neural operators (FNOs) can learn highly nonlinear mappings between function spaces, and have recently become a popular tool for learning responses of complex physical systems. However, to achieve good accuracy and efficiency, FNOs rely on the Fast Fourier transform (FFT), which is restricted to modeling problems on rectangular domains. To lift such a restriction and permit FFT on irregular geometries as well as topology changes, we introduce domain agnostic Fourier neural operator (DAFNO), a novel neural operator architecture for learning surrogates with irregular geometries and evolving domains. The key idea is to incorporate a smoothed characteristic function in the integral layer architecture of FNOs, and leverage FFT to achieve rapid computations, in such a way that the geometric information is explicitly encoded in the architecture. In our empirical evaluation, DAFNO has achieved state-of-the-art accuracy as compared to baseline neural operator models on two benchmark datasets of material modeling and airfoil simulation. To further demonstrate the capability and generalizability of DAFNO in handling complex domains with topology changes, we consider a brittle material fracture evolution problem. With only one training crack simulation sample, DAFNO has achieved generalizability to unseen loading scenarios and substantially different crack patterns from the trained scenario. Our code and data accompanying this paper are available at \url{https://github.com/ningliu-iga/DAFNO}. 
\end{abstract}

\section{Introduction}
%\vspace{-1mm}

Deep learning surrogate models provide a useful data-driven paradigm to accelerate the PDE-solving and calibration process of scientific modeling and computing problems. Among others, a wide range of scientific computing applications entail the learning of solution operators, i.e., the learning of infinite dimensional function mappings between any parametric dependence to the solution field. A prototypical instance is the case of solving Navier-Stokes equations in fluid mechanics, where the initial input needs to be mapped to a temporal sequence of nonlinear parametric solutions. The demand for operator learning has sparked the development of neural operator based methods \citep{li2020neural,li2020multipole,li2020fourier,you2022nonlocal,you2022physics,you2022learning,goswami2022physics,liu2022ino,gupta2021multiwavelet,lu2019deeponet,cao2021choose,hao2023gnot,li2022transformer,yin2022continuous}, with one of the most popular architectures being Fourier Neural Operators (FNOs) \citep{li2020fourier,you2022learning}.

The success of FNOs can be mostly attributed to its convolution-based integral kernel that learns in a resolution-invariant manner and the computationally efficient evaluation achieved via Fast Fourier Transform (FFT) \citep{brigham1988fast}. While learning in the spectral domain is fast, the latter comes at a cost: the computational domain of the underlying problem needs to be rectangular with uniformly meshed grids. This is often intractable as the domain of interest is, more often than not, irregular. An often taken trick for applying FNO to irregular domains is to embed the original domain into a larger rectangular domain and zero-pad or extrapolate on the redundant space \citep{lu2022comprehensive}. This poses two potential problems, one being the possible numerical errors and even instabilities due to the discontinuity at the original domain boundary (e.g., the Gibbs phenomenon \citep{gottlieb1997gibbs}) and the other, perhaps
more importantly, being the fact that the padding/extrapolating techniques cannot handle domains with shallow gaps, as is the case in object contact and crack propagation problems. 
%and the other, perhaps more importantly, being the fact that the model is unaware of the original domain boundary. 
Meanwhile, another line of work emphasizes the learning of a diffeomorphic mapping between the original domain and a latent domain with uniform grids on which FNO can be applied \citep{li2022fourier}. However, in this approach the changes on the boundary and the domain topology can only be informed via the learned diffeomorphism, which results in approximation errors when tested on a new domain geometry and possible failure when a change in topology is involved on the domain geometry.

In this work, we aim to design FNO architectures that explicitly embed the boundary information of irregular domains, which we coin Domain Agnostic Fourier Neural Operator (DAFNO). This is inspired by the recent work in convolution-based peridynamics \citep{jafarzadeh2022general} in which bounded domains of arbitrary shapes are explicitly encoded in the nonlocal integral formulation. We argue that, by explicitly embedding the domain boundary information into the model architecture, DAFNO is able to learn the underlying physics more accurately, and the learnt model is generalizable to changes on the domain geometry and topology. Concretely, we construct two practical DAFNO variants, namely, eDAFNO that inherits the explicit FNO architecture \citep{li2022fourier} and iDAFNO that is built upon the implicit FNO (IFNO) architecture characterizing layer-independent kernels \citep{you2022learning}. Moreover, a boundary smoothening technique is also proposed to resolve the Gibbs phenomenon and retain the fidelity of the domain boundary. In summary, the primary contributions of the current work are as follows:\vspace{-1mm}
\begin{itemize}
\item{We propose DAFNO, a novel Fourier neural operator architecture that explicitly encodes the boundary information of irregular domains into the model architecture, so that the learned operator is aware of the domain boundary, and generalizable to different domains of complicated geometries and topologies.\vspace{-1mm}}
\item{By incorporating a (smoothened) domain characteristic function into the integral layer, our formulation resembles a nonlocal model, such that the layer update acts as collecting interactions between material points inside the domain and cuts the non-physical influence outside the domain. As such, the model preserves the fidelity of the domain boundary as well as the convolution form of the kernel that retains the computational efficiency of FFT.\vspace{-1mm}}
\item{We demonstrate the expressivity and generalizability of DAFNO across a wide range of scientific problems including constitutive modeling of hyperelastic materials, airfoil design, and crack propagation in brittle fracture, and show that our learned operator can handle not only irregular domains but also topology changes over the evolution of the solution.}
\end{itemize}

\section{Background and related work}
\vspace{-1.1mm}

The goal of this work is to construct a neural network architecture to learn common physical models on various domains. Formally, given $\mcD:=\{(\gb_i|_{\omg_i},\ub_i|_{\omg_i})\}_{i=1}^N$, a labelled set of function pair observations both defined on the domain $\xb\in\omg_i\subset\real^s$. We assume that the input $\{\gb_i(\xb)\}$ is a set of independent and identically distributed (i.i.d.) random fields from a known probability distribution $\mu$ on $\mcA(\real^{d_g})$, a Banach space of functions taking values in $\real^{d_g}$. $\ub_i(\xb)\in\mcU(\real^{d_u})$, possibly noisy, is the observed corresponding response taking values in $\real^{d_u}$. Taking mechanical response modeling problem for example, $\omg_i$ is the shape of the object of interest, $\gb_i(\xb)$ may represent the boundary, initial, or loading conditions, and $\ub_i(\xb)$ can be the resulting velocity, pressure, or displacement field of the object. We assume that all observations can be modeled by a common and possibly unknown governing law, e.g., balance laws, and our goal is to construct a surrogate operator mapping,  $\tilde{\mcG}$, from $\mcA$ to $\mcU$ such that%\vspace{-2mm}
\begin{equation}\label{eqn:pde_single}
\tilde{\mcG}[\gb_i;\theta](\xb)\approx \ub_i(\xb),\;\forall \xb\in\omg_i.%\vspace{-2mm}
\end{equation}
Here, $\theta$ represents the (trainable) network parameter set.

In real-world applications, the domain $\omg_i$ can possess different topologies, e.g., in contact problems \citep{Benson2004,Simo1992} and material fragmentation problems \citep{LuyBen11,agwai2011predicting,silling2003dynamic}, and/or evolve with time, as is the case in large-deformation problems \citep{Shadden:2010ca} and fluid--structure interaction applications \citep{KuHuDe03,kamensky2017immersogeometric}. Hence, it is desired to develop an architecture with  generalizability across various domains of complex shapes, so that the knowledge obtained from one geometry can be transferable to other geometries.

% with a parametric PDE form
% \begin{equation}\label{eqn:pde_single}
% \begin{aligned} 
% \mcK [\ub_i] (\xb)=\gb_i (\xb),\quad&\xb\in \omg.
% \end{aligned}
% \end{equation}
% $\mcK$ is the operator representing the possibly unknown governing law, e.g., balance laws.

\vspace{-0.5mm}

\subsection{Learning solution operators of hidden physics}
\vspace{-0.5mm}

In real-world physical problems, predicting and monitoring complex system responses are ubiquitous in many applications. 
%For many decades, physics-based PDEs have been commonly employed for predicting and monitoring complex system responses, then 
For these purposes, physics-based PDEs and tranditional numerical methods have been commonly employed. However, %three fundamental challenges present. Firstly, the particular choice of governing PDE laws is often determined \textit{a priori}, and free parameters are often tuned to reach good agreements with observed data, making the rigorous calibration and validation process challenging. Secondly, 
traditional numerical methods are solved for specific boundary,  initial, and loading conditions $\gb$ on a specific domain $\omg$. Hence, the solutions are not generalizable to other domains and operating conditions. %Moreover, many classical PDE solvers also rely on a preprocessing step of mesh generation, which makes them cumbersome in complex physical applications involving changes in computational domains. %, as is the case in fluid--structure interaction \cite{KuHuDe03,kamensky2017immersogeometric} and material fragmentation problems \cite{LuyBen11,agwai2011predicting,silling2003dynamic}.
%Especially for complex PDE systems such as turbulence flows and heterogeneous materials modeling problems, a very fine discretization is usually required, which is time-consuming.

To provide a more efficient and flexible surrogate model for physical response prediction, %several recent efforts have been devoted to exploring machine learning based solutions. In particular, 
there has been significant progress in the development of deep neural networks (NNs) and scientific machine learning models \citep{ghaboussi1998autoprogressive,ghaboussi1991knowledge,carleo2019machine,karniadakis2021physics,zhang2018deep,cai2022physics,pfau2020ab,he2021manifold,besnard2006finite}. Among others, neural operators \citep{li2020neural,li2020multipole,li2020fourier,you2022nonlocal,Ong2022,gupta2021multiwavelet,lu2019deeponet,lu2021learning,goswami2022physics,tripura2022wavelet} show particular promise in learning physics of complex systems: %. 
% In contrast to classical neural networks which maps between finite-dimensional vectors, 
%Neural operators aim to learn maps between inputs of a dynamical system and its state, so that the network can serve as a surrogate for a solution operator 
compared with classical NNs, neural operators are resolution independent and generalizable to different input instances.
Therefore, once the neural operator is trained, solving for a new instance of the boundary/initial/loading condition with a different discretization only requires a forward pass. %Hence, the resultant solver is 
%Moreover, comparing with the classical PDE modeling approaches, neural operators require only data with no knowledge of the underlying PDE. All 
These advantages make neural operators a useful tool to many physics modeling problems \citep{yin2022simulating,goswami2022physics,yin2022interfacing, li2020neural,li2020multipole,li2020fourier,lu2022comprehensive,lu2021one}. 

\vspace{-0.5mm}

\subsection{Neural operator learning}
\vspace{-0.5mm}

Here, we first introduce %the Fourier neural operator model \cite{li2020fourier} and 
the basic architecture of the general integral neural operators \citep{li2020neural,li2020multipole,li2020fourier,you2022nonlocal,you2022learning}, 
%converts the convolution operation in the neural network into a multiplication operation through discrete Fourier transform, which can greatly improve the computational efficiency.
which are comprised of three building blocks. First, the input function, $\gb(\xb)\in\mcA$, is lifted to a higher-dimensional representation via $\hb^0(\xb)=\mathcal{P}[\gb](\xb):=P[\xb,\gb(\xb)]^T+\pb$, where $P\in\real^{(s+d_g)\times d_h}$ and $\pb\in\real^{d_h}$ define an affine pointwise mapping. Then, the feature vector function $\hb^0(\xb)$ goes through an iterative layer block, where the layer update is defined via the sum of a local linear operator, a nonlocal integral kernel operator, and a bias function:  $\hb^{l+1}(\xb)=\mathcal{J}^{l+1}[\hb^l](\xb)$. Here, $\hb^l(\xb)\in\real^{d_h}$, $l=0,\cdots,L$, is a sequence of functions representing the values of the network at each hidden layer. $\mathcal{J}^1,\cdots,\mathcal{J}^{L}$ are the nonlinear operator layers defined by the particular choice of networks. Finally, the output $\ub(\xb)\in\mcU$ is obtained via a projection layer by mapping the last hidden layer representation $\hb^L(\xb)$ onto $\mcU$ as:
$\ub(\xb)=\mathcal{Q}[\hb^L](\xb):=Q_2\sigma(Q_1\hb^L(\xb)+\qb_1)+\qb_2$. $Q_1$, $Q_2$, $\qb_1$ and $\qb_2$ are the appropriately sized matrices and vectors that are part of the learnable parameter set, and $\sigma$ is an activation function (e.g., ReLU \citep{he2018relu} or GeLU).

%The neural operator can be employed to learn an approximation for the solution operator, $\mcG$. 
Then, the system response can be learnt by constructing a surrogate operator of \eqref{eqn:pde_single}: $\tilde{\mcG}[\gb;\theta](\xb):=\mathcal{Q}\circ\mathcal{J}^1\circ\cdots\circ\mathcal{J}^L\circ\mathcal{P}[\gb](\xb)\approx \ub(\xb)$, by solving the network parameter set $\theta$ via an optimization problem: \vspace{-2mm}
\begin{equation}\label{eqn:opt}
\min_{\theta\in\Theta}\mcL_{\mcD}(\theta):=%\min_{\theta\in\Theta}\mathbb{E}_{\gb\sim\mu}[C(\tilde{\mcG}[\gb;\theta],\mcG[\gb])]\approx 
\min_{\theta\in\Theta}\sum_{i=1}^N[C(\tilde{\mcG}[\gb_i;\theta],\ub_i)] \text{ .}\vspace{-1mm}
\end{equation}
Here, $C$ denotes a properly defined cost functional (e.g., the relative mean square error) on $\omg_i$. %definedWe also point out that, in contrast to classical PDE-based approaches, neural operators can be trained directly from data, and hence their applications are not restricted to PDE solving problems.

%and  $\mcD:=\{\gb_i(\xb),\ub_{i}(\xb)\}_{i=1}^{N}$ denotes the support dataset containing available loading field/response field data pairs.
%In particular,  $\mcG[\cdot,\cdot;\theta]:(\fb,\ub^D)\rightarrow \ub$ was constructed, that delivers solutions of the system for any input $\ub^D$ and $\fb$. Here $\theta$ is the set of parameters in the network architecture to be inferred by fitting the experimental measurements data $\mcD=\{\ub_i(\xb),\ub^D_i(\xb),\fb_i(\xb)\}_{i=1}^N$.
%To achieve this goal, we plan to parametrize the mapping $\mcG$ as integral operators, 
%Differs from the classical neural network (NN) approaches where the solution operator is parameterized between finite-dimensional Euclidean spaces, in $\mcG$, the set of nodes were treated as a continuum so that the value of the network at each layer is a continuous function of a "space" variable. 
%to map the last feature layer $\hb(\cdot,L)$ onto the output $\ub(\cdot)$. %Then, the model writes: 
%\begin{equation}\label{eqn:ino}
%\mcG[\fb,\ub^D;\theta](\xb):=\mathcal{Q}\circ\mathcal{J}^L\circ \mathcal{P}[\fb,\ub^D](\xb)\approx \ub(\xb).
%\end{equation}

\vspace{-1mm}
\subsection{Fourier neural operators}
\vspace{-1mm}

The Fourier neural operator (FNO) is first proposed in \citet{li2020fourier} with its iterative layer architecture given by a convolution operator:\vspace{-2mm}
\begin{align}
\mathcal{J}^{l}[\hb](\xb):=&\sigma\biggl(W^l\hb(\xb)+\cb^l + \int_\omg \kappa(\xb-\yb;\vb^l)\hb(\yb) d\yb\biggr)\text{ ,}\label{eqn:fno0}\vspace{-1mm}
\end{align}
where $W^l\in\real^{d_h\times d_h}$ and $\cb^l\in\real^{d_h}$ are learnable tensors at the $l$-th layer, and $\kappa\in\real^{d_h\times d_h}$ is a tensor kernel function with parameters $\vb^l$. When a rectangular domain $\omg$ with uniform meshes is considered, the above convolution operation can be converted to a multiplication operation through discrete Fourier transform:
\begin{align}
\nonumber\mathcal{J}^{l}[\hb](\xb)=&\sigma\left(W^l\hb(\xb)+ \cb^l+\mathcal{F}^{-1}[\mathcal{F}[\kappa(\cdot;\vb^l)]\cdot \mathcal{F}[\hb(\cdot)]](\xb)\right)\text{ ,}\label{eq:fno}
\end{align}
where $\mathcal{F}$ and $\mathcal{F}^{-1}$ denote the Fourier transform and its inverse, respectively, which are computed using the FFT algorithm to each component of $\hb$ separately. The FFT calculations greatly improve the computational efficiency due to their quasilinear time complexity, but they also restrict the vanilla FNO architecture to rectangular domains $\omg$ \citep{lu2022comprehensive}.

To enhance the flexibility of FNO in modeling complex geometries, in \citet{lu2022comprehensive} the authors proposed to pad and/or extrapolate the input and output functions into a larger rectangular domain. However, such padding/extrapolating techniques are prone to numerical instabilities \citep{gottlieb1997gibbs}, especially when the domain is concave and/or with complicated boundaries. As shown in \citet{lu2022comprehensive}, the performance of dgFNO+ substantially deteriorates when handling complicated domains with notches and gaps. In Geo-FNO \citep{li2022fourier}, an additional neural network is employed and trained from data, to continuously map irregular domains onto a latent space of rectangular domains. As a result, the vanilla FNO can be employed on the rectangular latent domain. However, this strategy relies on the continuous mapping from the physical domain to a rectangular domain, hence it is restricted to relatively simple geometries with no topology change.

\section{Domain Agnostic Fourier Neural Operators}

\vspace{-1mm}

\begin{figure}[!t]\centering
\includegraphics[width=.6\linewidth]{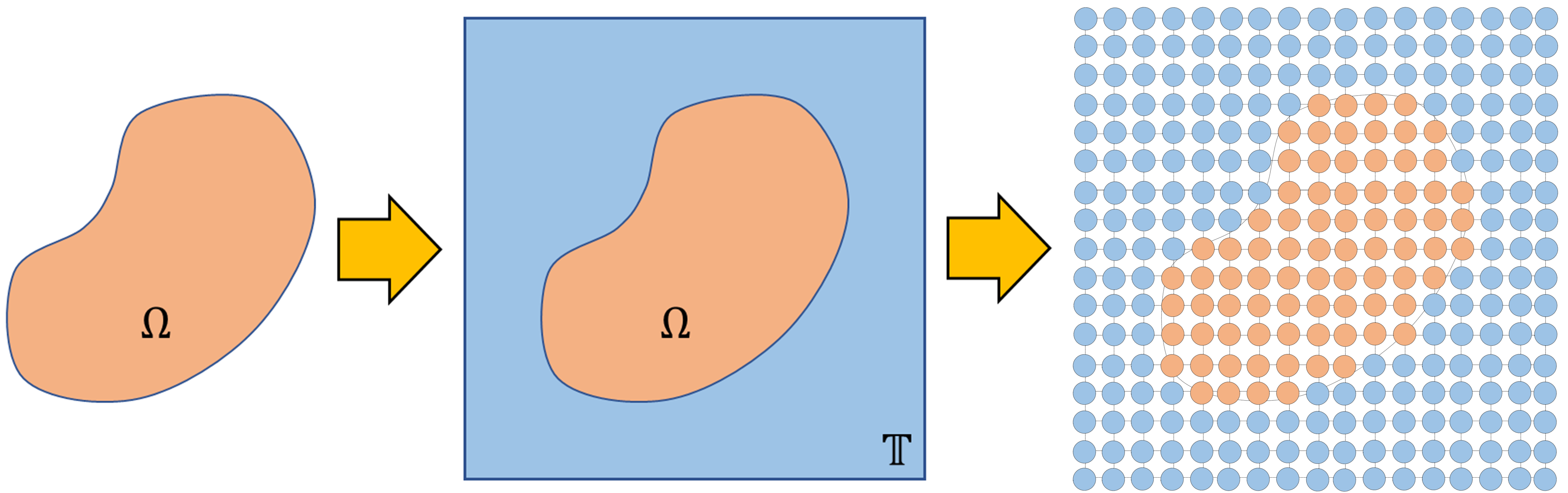}\vspace{-1mm}
 \caption{A schematic of extending an arbitrarily shaped domain $\Omega$ to a periodic box $\mathbb{T}$ and its discretized form in 2D.}
 \label{fig:domain_extension}\vspace{-1mm}
\end{figure}

In this section, we introduce Domain Agnostic Fourier Neural Operator (DAFNO), which features the generalizability %learns an operator applicable to 
%the full generalizability of FNOs
%where we extend the standard FNO structure \cite{li2020fourier}
to new and unseen domains of arbitrary shapes and different topologies. The key idea % evolution of bounded domains. 
is to explicitly encode the domain information in the design %of the FNO architecture 
while retaining the convolutional architecture in the iterative layer of FNOs. In what follows, we present the eDAFNO architecture based on the standard/explicit FNO model \citep{li2020fourier}, while the iDAFNO architecture based on the implicit FNO model \citep{you2022learning} is provided in Appendix~\ref{app:a}.
%To this aim, inspired by the fast convolution-based peridynamics \cite{jafarzadeh2022general} which is an FFT-based numerical method developed for efficient nonlocal computation, we explicitly embed the domain of interest into the DAFNO architecture.

Concretely, %let $\Omega(t)$ denote an arbitrarily shaped and evolving domain at time $t$. We 
we enclose the physical domain of interest, $\Omega$, by a (slightly larger) periodic box $\mathbb{T}$, as shown in Figure~\ref{fig:domain_extension}. 
%Recall the integral form of a FNO layer:
%\begin{align}\label{eq:fno_layer}
%\bm{h}^{l+1}_{\bm{x}, t} =& \sigma \left(\int_\Omega \kappa(\bm{x}-\bm{y};\bm{v}_l)\bm{h}^l_{\bm{y},t}d\bm{y} + W_l \bm{h}^l_{\bm{x},t}\right) \text{ ,}
%\end{align}
Next, we define the following domain characteristic function:%\vspace{-1mm}
\begin{equation}
  \chi(\bm{x}) =
    \begin{cases}
      1 & \bm{x} \in \Omega\\
      0 & \bm{x} \in \mathbb{T}\setminus\Omega%\vspace{-1mm}
    \end{cases},
\end{equation}
which encodes the domain information of different geometries. 
% which can also be written in the nonlocal Laplacian operator sense:
% \begin{align}
% \bm{h}^{l+1}_{\bm{x}, t} =& \sigma \left(\int_\Omega \kappa(\bm{x}-\bm{y};\bm{v}_l)(\bm{h}^l_{\bm{y},t}-\bm{h}^l_{\bm{x},t})d\bm{y} + W_l \bm{h}^l_{\bm{x},t}\right) \text{ .}
% \end{align}
% We choose to work with the nonlocal version in this study, but the same procedure can be easily applied to Eq.~\eqref{eq:fno_layer}.
Inspired by \citet{jafarzadeh2022general}, we incorporate the above-encoded domain information into the FNO architecture of \eqref{eqn:fno0}, by multiplying the integrand in its convolution integral with $\chi(\bm{x})\chi(\bm{y})$:%\vspace{-1mm}
\begin{align}
\begin{split}
\mathcal{J}^{l}[\hb] =& \sigma \left(\int_{\mathbb{T}} \chi(\bm{x})\chi(\bm{y}) \kappa(\bm{x}-\bm{y};\bm{v}^l)(\bm{h}(\bm{y})-\bm{h}(\bm{x}))d\bm{y}  + W^l \bm{h}(\bm{x})+\cb^l\vphantom{\int_\Omega}\right) \text{ .}\label{eq:no_with_chi} %\vspace{-1mm}
\raisetag{20pt}
\end{split}
\end{align}
Herein, we have followed the practice in \citet{you2022nonlocal} and reformulated \eqref{eqn:fno0} to a nonlocal Laplacian operator, which is found to improve training efficacy. %We point out that the same procedure can also be applied to the original integral operator in \eqref{eqn:fno0}. 
By introducing the term $\chi(\bm{x})\chi(\bm{y})$, the integrand vanishes when either point $\bm{x}$ or $\bm{y}$ is positioned inside $\Omega$ and the other is positioned outside. This modification eliminates any undesired interaction between the regions inside and outside of $\Omega$. As a result, it tailors the integral operator to act on $\Omega$ independently and is able to handle different domains and topologies. With this modification, the FFT remains applicable, since the convolutional structure of the integral is preserved and the domain of operation yet spans to the whole rectangular box $\mathbb{T}$. In this context, \eqref{eq:no_with_chi} can be re-organized as:%\vspace{-1mm}
\begin{align*}
\mathcal{J}^{l}[\hb] =& \sigma \left(\chi(\bm{x})\bigl(\int_{\mathbb{T}} \kappa(\bm{x}-\bm{y};\bm{v}^l)\chi(\bm{y})\bm{h}(\bm{y})d\bm{y}- \bm{h}(\bm{x}) \int_{\mathbb{T}} \kappa(\bm{x}-\bm{y};\bm{v}^l)\chi(\bm{y})d\bm{y} + W^l \bm{h}(\bm{x})+\cb^l\bigr)\vphantom{\int_\Omega}\right).%\vspace{-1mm}
\end{align*}
%where the second integral can be merged into $W_l \bm{h}_l(\bm{x},t)$ and form the DAFNO version of Eq.~\eqref{eq:fno_layer}, or kept explicitly if desired. In this study, we keep this term in the architecture. 
Note that multiplying $W^l \bm{h}(\bm{x})+\cb^l$ with $\chi(\bm{x})$ does not alter the computational domain. Now that the integration region is a rectangular box $\mathbb{T}$, the FFT algorithm and its inverse can be readily applied, and hence we can further express the eDAFNO architecture as:%\vspace{-1mm}
\begin{align}
\begin{split}
\mathcal{J}^{l}[\hb] :=& \sigma \biggl(\chi(\bm{x}) \bigl(\mathcal{I}(\chi(\cdot) \bm{h}(\cdot);\vb^l)  -\bm{h}(\bm{x}) \mathcal{I}(\chi(\cdot);\vb^l) + W^l \bm{h}(\bm{x})+\cb^l \bigr) \biggr)\text{ ,}\\
\text{where }\;\; & \mathcal{I}(\circ;\vb^l) := \mathcal{F}^{-1} \bigl[\mathcal{F}[\kappa(\cdot;\vb^l)] \cdot \mathcal{F}[\circ]\bigr]\text{ .}
\label{eq:fno_with_chi}%\vspace{-1mm}
\raisetag{14pt}
\end{split}
\end{align}
%where the time variable $t$ can be dropped in case of stationary/static problems.

\begin{figure}[!t]\centering
    \includegraphics[width=.7\linewidth]{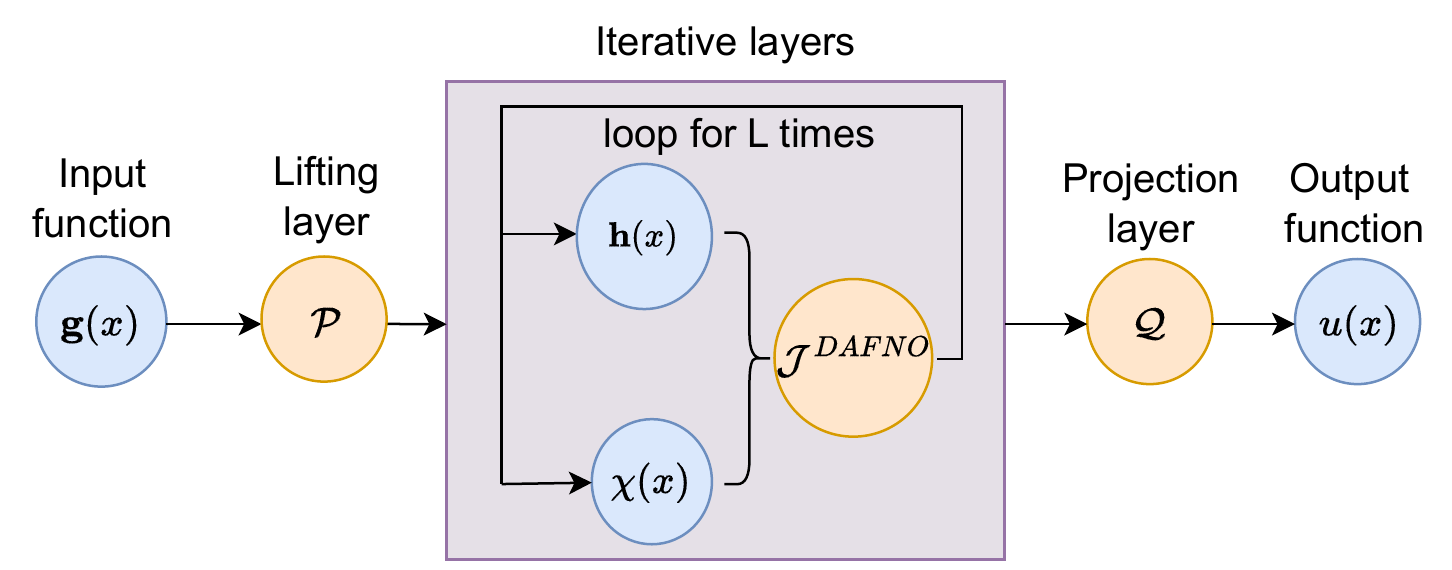}
    \caption{An illustration of the proposed DAFNO architecture. We start from the input function $\gb(x)$. After lifting, the iterative Fourier layers are built that explicitly embed the encoded domain information, $\chi$. Lastly, we project the last hidden layer representation to the target function space.\vspace{-1mm}}
\label{fig:dafno_architecture}
\end{figure}

An illustration of the DAFNO atchitecture is provided in Figure~\ref{fig:dafno_architecture}. Note that this architectural modification is performed at the continuum level and therefore is independent of the discretization. Then, the box $\mathbb{T}$ can be discretized with structured grids (cf. Figure~\ref{fig:domain_extension}), as is the case in standard FNO. 
%The discrete characteristic function is either 0 or 1 for each node, depending on whether the node is positioned inside or outside of $\Omega$.

\begin{figure}[!h]\centering
\includegraphics[width=1.0\linewidth]{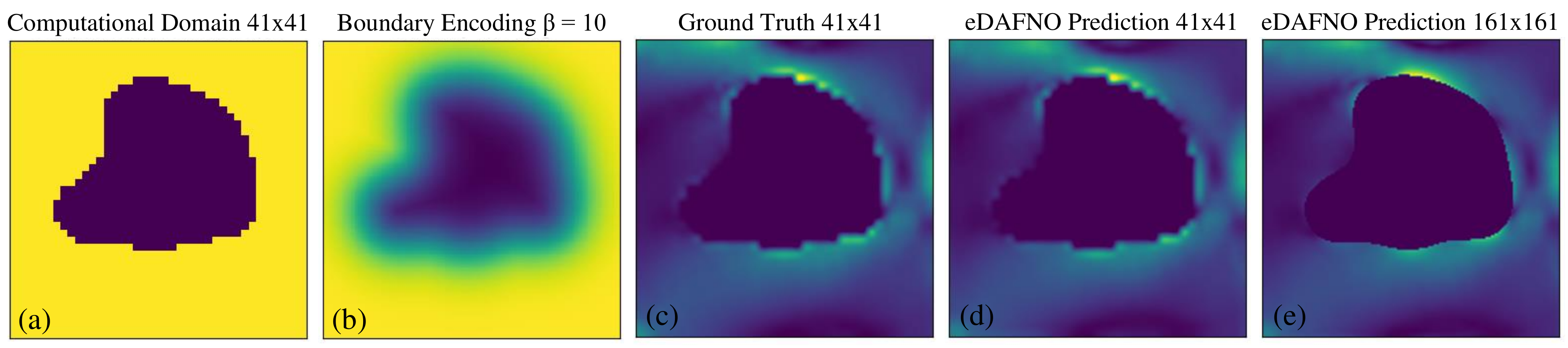}\vspace{-1mm}
 \caption{An illustration on a hyperelasticity sample: (a) sharp characteristic function $\chi$, (b) smoothed characteristic function $\chi$, (c) ground truth, (d) eDAFNO (trained using 41$\times$41 discretization) prediction from the same resolution, and (e) zero-shot super-resolution prediction from eDAFNO (trained using 41$\times$41 discretization and evaluated directly on 161$\times$161 discretization).\vspace{-2mm}}
 \label{fig:super_resolution}
\end{figure}

Although the proposed DAFNO architecture in \eqref{eq:fno_with_chi} can handle complex generalization in domains, it has a potential pitfall: since the characteristic function is not continuous on the domain boundary, its Fourier series cannot converge uniformly and the FFT result would present fictitious wiggling near the discontinuities (i.e., the Gibbs phenomenon \citep{day1965developments}). As a consequence, the introduction of $\chi$ can potentially jeopardize the computational accuracy. To improve the efficacy of DAFNO, we propose to replace the sharp characteristic function, $\chi$, with a smoothed formulation:
\begin{equation}\label{eqn:smooth_chi}
  \tilde{\chi}(\bm{x}) := \text{tanh}(\beta\text{dist}(\xb,\partial\omg))(\chi(\xb)-0.5)+0.5 \text{ .}
\end{equation}
Here, the hyperbolic tangent function $\tanh(z):=\frac{\exp(z)-\exp(-z)}{\exp(z)+\exp(-z)}$, $\text{dist}(\xb,\partial\omg)$ denotes the (approximated) distance between $\xb$ and the boundary of domain $\omg$, and $\beta$ controls the level of smoothness, which is treated as a tunable hyperparameter. An illustration of the effect of the smoothed $\tilde{\chi}$ is displayed in Figure~\ref{fig:super_resolution}, with additional plots and prediction results with respect to different levels of smoothness, $\beta$, provided in Appendix~\ref{app:b1}. In what follows, the $\tilde{\cdot}$ sign is neglected for brevity.

\textbf{Remark:} We point out that the proposed smoothed geometry encoding technique, although simple, is substantially different from existing function padding/extrapolation techniques proposed in \cite{lu2022comprehensive} who cannot handle singular boundaries (as in the airfoil tail of our example 2) nor notches/shallow gaps in the domain (as in the crack propagation of our example 3). Our proposed architecture in \eqref{eq:no_with_chi} is also more sophiscated than a trivial zero padding at each layer in that the characteristic function $\chi(\xb)$ is multiplied with the integrand, whereas the latter breaks the convolutional structure and hinders the application of FFTs.

% \begin{figure}[!t]\centering
% \includegraphics[width=.99\linewidth]{figures/boundary_encoding.pdf}
%  \caption{An illustration of the effect of the smoothing coefficient on the resulting boundary encoding. The larger the smoothing level $\beta$ is, the sharper and narrower the encoded boundary becomes.}
%  \label{fig:boundary_encoding}
% \end{figure}

%\vspace{-1mm}
\section{Numerical examples}\label{sec:exp}
\vspace{-1mm}

In this section, we demonstrate the accuracy and expressivity of DAFNO across a wide variety of scientific problems. We compare the performance of DAFNO against other relevant scientific machine learning models, %that can be served as surrogates, 
including FNO \citep{li2020fourier}, Geo-FNO \citep{li2022fourier}, IFNO \citep{you2022learning}, F-FNO \citep{tran2022factorized}, GNO \citep{li2020neural}, DeepONet \citep{lu2019deeponet}, and UNet \citep{ronneberger2015u}. In particular, we carry out three experiments on irregular domains, namely, constitutive modeling of hyperelasticity in material science, airfoil design in fluid mechanics, and crack propagation with topology change in fracture mechanics. For fair comparison, the hyperparameters of each model are tuned to minimize the error on validation datasets, including initial learning rate, decay rate for every 100 epochs, smoothing parameter, and regularization parameter, while the total number of epochs is restricted to 500 for computational efficiency. The relative L2 error is reported as comparison metrics for both the training and test datasets.  The experiments of each method are repeated on 5 trials with 5 different random seeds, and the mean and standard deviation of the errors are reported. Further details and additional results are provided in Appendix~\ref{app:b}.

%\vspace{-1mm}
\subsection{Constitutive modeling of hyperelasticity}
\vspace{-1mm}

We start with a hyperelasticity problem in material science that models the fundamental principle of constitutive relations. The high-fidelity synthetic data in this benchmark is governed by% the following equation:
\begin{equation}\label{eq:elas}
\rho \frac{\partial^2 \bm{u}}{\partial t^2} + \nabla \cdot \bm{\sigma} = 0 \text{ ,}
\end{equation}
where $\rho$ denotes the mass density, $\bm{u}$ and $\bm{\sigma}$ represent the corresponding displacement and stress fields, respectively. The computational domain is enclosed by a unit cell $[0, 1]^2$ of which the center exists a randomly shaped void, as described by its radius $r=0.2+\frac{0.2}{1+exp(\tilde{r})}$ and $\tilde{r}\sim\mathbb{N}(0, 4^2(-\nabla+3^2)^{-1})$. The bottom edge is fixed and the top edge is subjected to a tensile traction of $\bm{t}=[0, 100]$. The underlying hyperelastic material is of the incompressible Rivlin-Saunders type. For training, we directly adopt the dataset in \citet{li2022fourier}, where a total of 1000, 200, 200 samples are selected for training, validation and testing, respectively. For this problem, the input is represented as point clouds and the target is the resulting stress field.

\begin{table}[!h]
    \caption{The total number of parameters (in millions) of selected models for hyperelasticity dataset.}
    \label{tab:elas_para}
    \centering
    {\small \centering
    \begin{tabular}{cccccccccc}
    \toprule
         model & eDAFNO & iDAFNO & FNO & IFNO & Geo-FNO & GNO & DeepONet & UNet & F-FNO\\
         \midrule
         nparam & 2.37 & 0.60 & 2.37 & 0.60 & 3.02 & 2.64 & 3.10 & 3.03 & 3.21\\
         \bottomrule
    \end{tabular}}
%\vskip -0.1in
\end{table}

\begin{table*}[h!]
\caption{Test errors for the hyperelasticity problem, where bold numbers highlight the best method.}% The experiments of each method are repeated on 5 trials with 5 different random seeds.}
\label{tab:elas_results}
%\vskip 0.15in
\begin{center}
{\small    \centering
%\begin{tabular}{|c|c|c|c|}
\begin{tabular}{ll|lll}
    \toprule
       \multicolumn{2}{c}{Model} & \multicolumn{3}{c}{\# of training samples} \\
        \cline{1-5}
       && 10 & 100 & 1000 \\
\hline
%eDAFNO, train & 6.800\%$\pm$0.670\% & 2.050\%$\pm$0.035\% & 0.664\%$\pm$0.014\%\\
\multirow{2}{*}{Proposed model}& eDAFNO & \bf{16.446}\%$\pm$\bf{0.472}\% & 4.247\%$\pm$0.066\% & \bf{1.094}\%$\pm$\bf{0.012}\%\\
%\hline
%iDAFNO, train & 7.266\%$\pm$0.923\% & 2.038\%$\pm$0.036\% & 0.812\%$\pm$0.012\%\\
&iDAFNO & 16.669\%$\pm$0.523\% & \bf{4.214}\%$\pm$\bf{0.058}\% & 1.207\%$\pm$0.006\%\\
\hline
\multirow{6}{*}{Baseline model}& FNO w/ mask & 
         19.487\%$\pm$0.633\%& 7.852\%$\pm$0.130\% & 4.550\%$\pm$0.062\%\\
&IFNO w/ mask & 19.262\%$\pm$0.376\%& 7.700\%$\pm$0.062\%& 4.481\%$\pm$0.022\%\\
&Geo-FNO & 
         28.725\%$\pm$2.600\%&10.343\%$\pm$4.446\% &2.316\%$\pm$0.283\%\\
&GNO& 29.305\%$\pm$0.321\% & 18.574\%$\pm$0.584\% & 13.007\%$\pm$0.729\%\\
&DeepONet& 35.334\%$\pm$0.179\% & 25.455\%$\pm$0.245\% & 11.998\%$\pm$0.786\%\\
&F-FNO& 35.672\%$\pm$3.852\% & 12.135\%$\pm$5.813\% & 3.193\%$\pm$1.622\%\\
&UNet& 98.167\%$\pm$0.236\% & 34.467\%$\pm$2.858\% & 5.462\%$\pm$0.048\%\\
\hline
\multirow{2}{*}{Ablation study} &FNO w/ smooth $\chi$ & 
         17.431\%$\pm$0.536\%&5.479\%$\pm$0.186\% & 1.415\%$\pm$0.025\%\\
%FNO w/ mask, train &         2.907\%$\pm$0.318\%&2.277\%$\pm$0.240\% &0.881\%$\pm$0.015\%\\
&IFNO w/ smooth $\chi$& 17.145\%$\pm$0.432\% & 5.088\%$\pm$0.146\%& 1.509\%$\pm$0.018\%\\
\bottomrule
\end{tabular}}
\end{center}
\vskip -0.1in
\end{table*}

% \paragraph{Ablation study}
\textbf{Ablation study } We first carry out an ablation study by comparing (a) the proposed two DAFNO models with (b) the baseline FNO/IFNO with the sharp characteristic function as input (denoted as FNO/IFNO w/ mask), and (c) the original FNO/IFNO with our proposed smoothened boundary characteristic function as input (denoted as FNO/IFNO w/ smooth $\chi$).
%the following four scenarios: (a) the original FNO with both the mask and the smoothened boundary encoding (denoted as smooth $\chi$) multiplied to the predictions separately, (b) the original IFNO with both the mask and smoothened chi multiplied to the predictions separately, (c) the proposed eDAFNO architecture, and (d) the proposed iDAFNO architecture. 
In this study, scenarios (b) and (c) aim to investigate the effects of our proposed boundary smoothing technique, and by comparing scenarios (a) and (c) we verify the effectiveness of encoding the boundary information in eDAFNO architecture. 
%Analogously, scenarios (b) and (d) are aimed to explore the robustness of the proposed boundary smoothing technique and the iDAFNO architecture separately. 
In addition, both FNO- and IFNO-based models are tested, with the purpose to evaluate the model expressivity when using layer-independent parameters in iterative layers. Three training dataset sizes (i.e., 10, 100, and 1000) are employed to explore the effect of the proposed algorithms on small, medium, and large datasets, respectively. The number of trainable parameters are reported in Table~\ref{tab:elas_para}. Following the common practice as in \cite{li2022fourier}, the hyperparameter choice of each model is selected by tuning the number of layers and the width (channel dimension) keeping the total number of parameters of the same magnitude.

The results of the ablation study are listed in Table~\ref{tab:elas_results}. Firstly, by directly comparing the results of FNO (and IFNO) with mask and with smooth boundary encoding, one can tell that the boundary smoothing technique helps to reduce the error. This is supported by the observation that FNO and IFNO with smooth $\chi$ consistently outperform their counterparts with mask in all data regimes, especially when a sufficient amount of data becomes available where a huge boost in accuracy can be achieved (by over 300\%). On the other hand, by encoding the geometry information into the iterative layer, %we compare DAFNO/iDAFNO against FNO/IFNO with smooth $\chi$, %as well as iDAFNO against IFNO with smooth $\chi$, 
the prediction accuracy is further improved, where eDAFNO and iDAFNO outperform FNO and IFNO with smoothed $\chi$ by 22.7\% and 20.0\% in large-data regime, respectively. Another interesting finding is that eDAFNO is 9.4\% more accurate compared to iDAFNO in the large-data regime, although only a quarter of the total number of parameters is needed in iDAFNO due to its layer-independent parameter setting. This effect is less pronounced as we reduce the amount of available data for training, where the performance of iDAFNO is similar to that of eDAFNO in the small- and medium-data regimes. This is because iDAFNO has a much smaller number of trainable parameters and therefore is less likely to overfit with small datasets. Given that the performance of eDAFNO and iDAFNO is comparable, it is our opinion that both architectures are useful in different applications. In Figure \ref{fig:super_resolution}, an example of the computational domain, the smoothened boundary encoding, the ground truth solution, and the eDAFNO prediction are demonstrated. To demonstrate the capability of prediction across resolutions, we train eDAFNO using data with 41$\times$41 grids then apply the model to provide prediction on 161$\times$161 grids--one can see that eDAFNO can generalize across different resolutions.

% \paragraph{Comparison against additional baselines}
\textbf{Comparison against additional baselines } We further compare the performance of DAFNO against additional relevant baselines, including GNO, Geo-FNO, F-FNO, DeepONet, and UNet. Note that the results of GNO, DeepONet, and UNet are obtained using the same settings as in \citet{li2022fourier}. Overall, the two DAFNO variants are significantly superior to other baselines in accuracy, with eDAFNO outperforming GNO, DeepONet, UNet, Geo-FNO, and F-FNO by 1088.9\%, 975.0\%, 399.3\%, 111.7\%, and 191.9\%, respectively, in large-data regime. DAFNO is also more memory efficient compared to Geo-FNO (the most accurate baseline), as it foregoes the need for additional coordinate deformation network. As shown in Table \ref{tab:elas_para}, when the layer-independent parameter setting in iDAFNO is taken into account, DAFNO surpasses Geo-FNO by 407.4\% in memory saving.

%\vspace{-1mm}
\subsection{Airfoil design}
\vspace{-2mm}

\begin{figure}[!h]\centering
\includegraphics[width=1.0\linewidth]{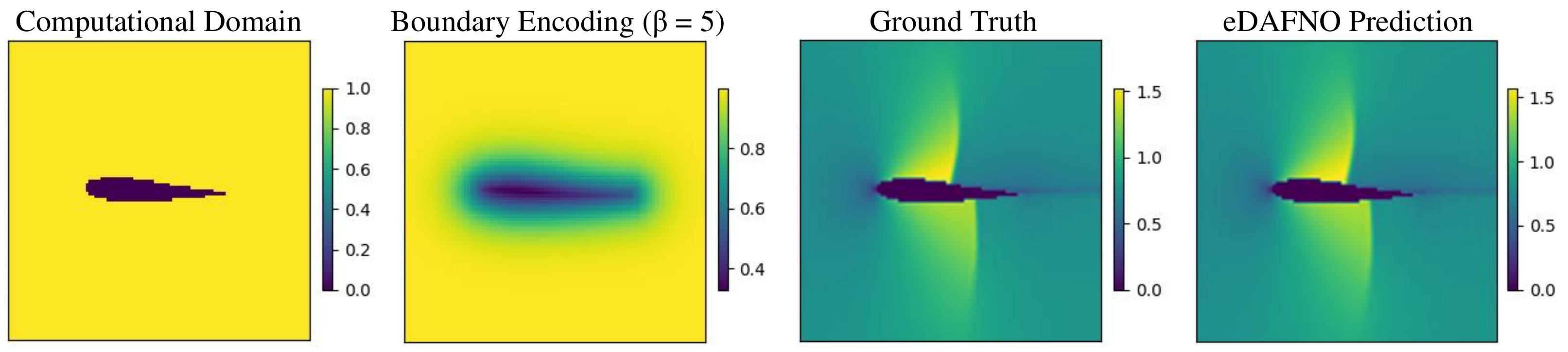}\vspace{-1mm}
 \caption{An illustration on a test sample from the airfoil design problem. From left to right: an illustration of the discretized computational domain, the smoothed boundary encoding (i.e., smoothed $\chi$), the ground truth, and the eDAFNO prediction.}
 \label{fig:airfoil}
\end{figure}

In this example, we investigate DAFNO's performance in learning transonic flow around various airfoil designs. Neglecting the effect of viscosity, the underlying physics can be described by the Euler equation:
\begin{equation}\label{eq:airfoil}
\frac{\partial \rho}{\partial t} + \nabla \cdot \left( \rho \bm{v}\right) = 0 \text{ ,} \quad
\frac{\partial \rho \bm{v}}{\partial t} + \nabla \cdot \left( \rho \bm{v} \otimes \bm{v} + p \mathbb{I}\right) = 0 \text{ ,} \quad
\frac{\partial E}{\partial t} + \nabla \cdot \left( \left( E+p\right) \bm{v}\right) = 0 \text{ ,}
\end{equation}
with $\rho$, $p$ and $E$ being the fluid density, pressure and the total energy, respectively, and $\bm{v}$ denoting the corresponding velocity field. The applied boundary conditions are: $\rho_\infty=1$, Mach number $M_\infty=0.8$, and $p_\infty=1$ on the far field, with no penetration enforced on the airfoil. The dataset used for training is directly taken from \citet{li2022fourier}, which consists of variations of the NACA-0012 airfoil and is divided into 1000, 100, 100 samples for training, validation and testing, respectively. For this problem, we aim to learn the resulting Mach number field based on a given mesh as input. An example of the computational domain, the smoothened boundary encoding, the ground truth, and the eDAFNO prediction is illustrated in Figure~\ref{fig:airfoil}.

We report in Table~\ref{tab:airfoil_results} our experimental observations using eDAFNO and iDAFNO, along with the comparison against FNO, Geo-FNO, F-FNO and UNet, whose models are directly attained from \citet{li2022fourier, tran2022factorized}. We can see that the proposed eDAFNO achieves the lowest error on the test dataset, beating the best result of non-DAFNO baselines by 24.9\% and Geo-FNO by 63.9\%, respectively. Additionally, iDAFNO reaches a similar level of accuracy with only a quarter of the total number of parameters employed, which is consistent with the findings in the previous example. With the properly trained DAFNO models, efficient optimization and inverse design are made possible.

\begin{table}[t!]
\caption{Results for the airfoil design problem, where bold numbers highlight the best method.\vspace{-1mm}}
\label{tab:airfoil_results}
\vskip -0.3in
\begin{center}
{\small    \centering
\begin{tabular}{ll|ll}
    \toprule
        \multicolumn{2}{c}{Model} & Train error & Test error\\
        \hline
         \multirow{2}{*}{Proposed model}&eDAFNO & 0.329\%$\pm$0.020\% & \bf{0.596}\%$\pm$\bf{0.005}\% \\
%\hline
         &iDAFNO & 0.448\%$\pm$0.012\% & 0.642\%$\pm$0.020\% \\

         &eDAFNO on irregular grids & 0.331\%$\pm$0.003\% & 0.659\%$\pm$0.007\% \\
\hline
         \multirow{3}{*}{Baseline model}&Geo-FNO & 1.565\%$\pm$0.180\% & 1.650\%$\pm$0.175\% \\
%\hline
         &F-FNO & 
         0.566\%$\pm$0.066\% & 0.794\%$\pm$0.025\% \\
%\hline
         &FNO w/ mask & 2.676\%$\pm$0.054\% & 3.725\%$\pm$0.108\% \\
%\hline
         &UNet w/ mask & 2.781\%$\pm$1.084\% & 4.957\%$\pm$0.059\% \\
\bottomrule
\end{tabular}}
\end{center}
\vskip -0.05in
\end{table}

\begin{figure}[!h]\centering
\includegraphics[width=1.0\linewidth]{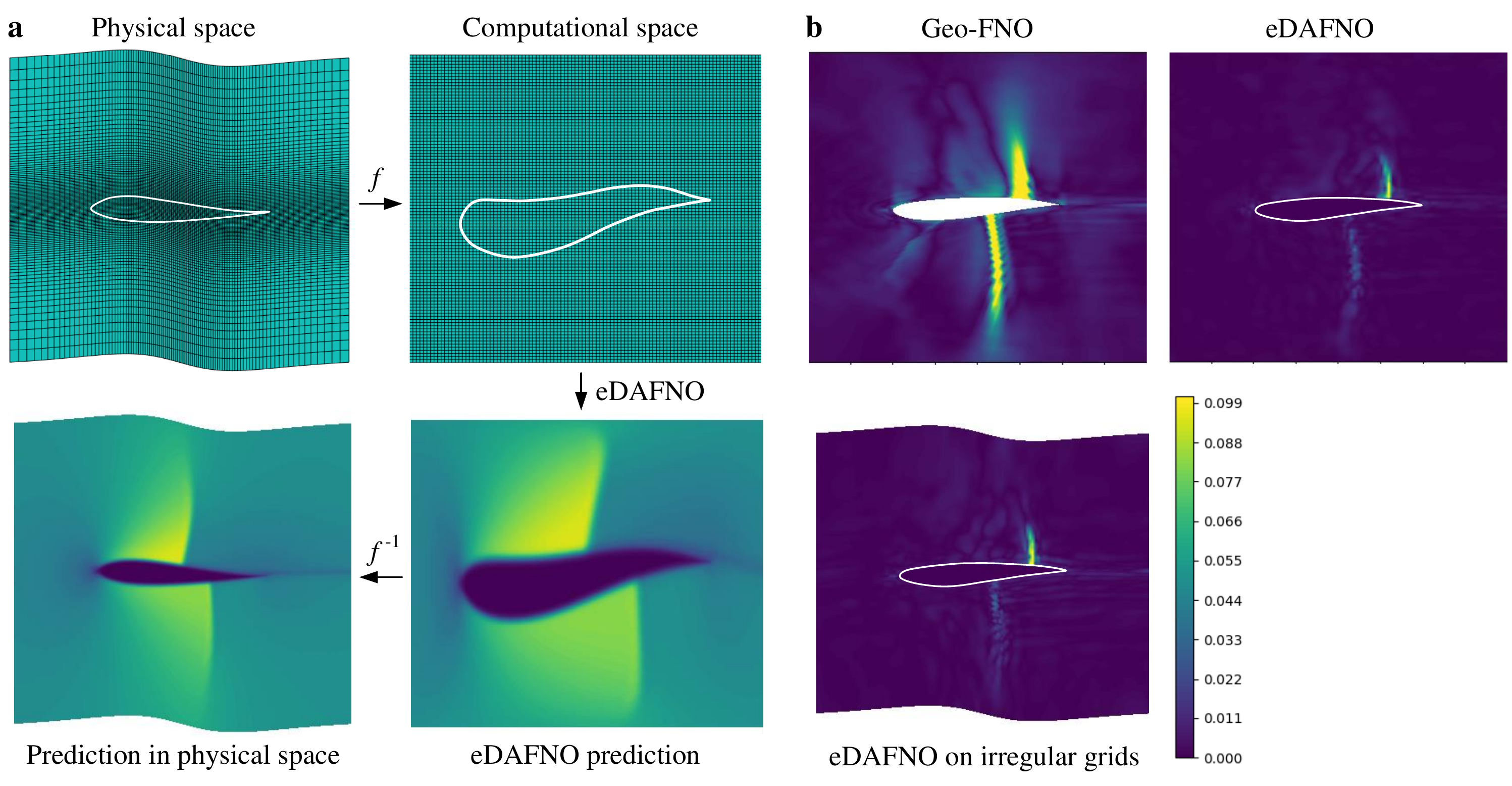}\vspace{-1mm}
 \caption{eDAFNO applied to the airfoil dataset on irregular grids. (a): the highly irregular and adaptive grids in the physical space is firstly deformed (via either an analytical mapping $f$ or a trainable neural network for grid deformation) to uniform grids in the computational space, on which eDAFNO is applied to learn the underlying physics. The learned prediction is then converted back to the physical space via the inverse mapping $f^{-1}$ or with another trainable neural network. (b): an illustration of the absolute error distribution of predictions in Geo-FNO, eDAFNO trained on uniform grids, and eDAFNO trained on irregular grids.}
 \label{fig:airfoil_grids_deformation}
 \vspace{-1mm}
\end{figure}

Aside from the enhanced predictability, DAFNO can also be easily combined with the grid mapping technique in Geo-FNO to handle non-uniform grids. In particular, no modification on the model architecture is required, where one just needs to include an analytical or trainable mapping from non-uniform/irregular mesh grids to uniform mesh grids. As a demonstration, we consider irregular grids of the airfoil dataset and use a pre-computed function to map irregular grids to regular grids. In the irregular grid set, we place more grid points near the airfoil to provide a better resolution near the important parts. The test error is provided in Table~\ref{tab:airfoil_results}, where the test loss using the eDAFNO learned model is $0.659\%\pm0.007\%$, which is similar to the DAFNO results on uniform grids. In Figure~\ref{fig:airfoil_grids_deformation}, we demonstrate the irregular mesh and the absolute error comparison across Geo-FNO, DAFNO on regular grids, and DAFNO on irregular grids. One can see that, while both DAFNOs substantially outperform Geo-FNO, the error contours from eDAFNO with an irregular mesh show a smaller miss-match region near the airfoil, illustrating the flexibility of DAFNO in meshing and its capability in resolving fine-grained features.

\subsection{Crack propagation with topology change in domain}
\vspace{-1mm}

In this example, we aim to showcase DAFNO’s capability in handling evolving domains by modeling crack propagation in brittle fracture. We emphasize that DAFNO represents the first neural operator that allows for learning with topology change. In the field of brittle fracture, a growing crack can be viewed as a change in topology, which corresponds to an evolving $\chi(t)$ in the DAFNO architecture. 
%In DAFNO, the topology evolves by updating $\chi$. 
In particular, we define the following time-dependent characteristic function:
\begin{equation}
  \chi(\bm{x},t) =
    \begin{cases}
      1 & \bm{x} \in \Omega(t)\\
      0 & \bm{x} \in \mathbb{T}\setminus\Omega(t)
    \end{cases},      
\end{equation}
where $\Omega(t)$ denotes the time-dependent domain/evolving topology. Employing time-dependent $\chi$ in \eqref{eq:fno_with_chi} keeps the neural operator informed about the evolving topology. In general, the topology evolution rule that determines $\Omega(t)$ can be obtained from a separate neural network or from physics as is the case in the current example. The high-fidelity synthetic data in this example is generated using 2D PeriFast software \citep{jafarzadeh2022perifast}, which employs a peridynamics (PD) theory for modeling fracture \citep{bobaru2016handbook}. A demonstration of the evolving topology, as well as further details regarding the governing equations and data generation strategies, is provided in Appendix~\ref{app:b3}.

In this context, we select eDAFNO as the surrogate model to learn the the internal force density operator. Specifically, let $u_1(\xb,t)$, $u_2(\xb,t)$ denote the two components of the displacement field $\ub$ at time $t$, and $L_1$, $L_2$ be the two components of the internal force density $\mathcal{L}[\ub]$. Given $u_1$, $u_2$ and $\chi$ as input data, we train two separate eDAFNO models to predict $L_1$ and $L_2$, respectively. %The trained eDAFNO models can then replace $\mathcal{L}$ in the high-fidelity solver. This means that any conventional second-order ODE solver can be used for time marching in the case of dynamic problems. 
Then, we substitute the trained surrogate models in the dynamic governing equation and adopt Velocity--Verlet time integration to update $\ub$ for the next time step, whereafter $\Omega$ and $\chi$ are updated accordingly.

The problem of interest is defined on a $40\text{ mm}\times40\text{ mm}$ thin plate with a pre-crack of length 10 mm at one edge, which is subjected to sudden, uniform, and constant tractions on the top and bottom edges. 
%Shock waves emanate from the edges, move toward the precrack, and initiate damage. 
Depending on the traction magnitude, crack grows at different speeds, may or may not bifurcate, and the final crack patterns can be of various shapes. %Figure~\ref{fig:initial_crack} illustrates the problem configuration.
% \begin{figure}[!t]\centering
% \includegraphics[width=.2\linewidth]{figures/initial_crack.pdf}
%  \caption{The crack propagation problem setup: a 2D plate with a pre-crack subjected to external tractions (denoted as $\sigma$ with a slight abuse of notation) on the top and bottom edges.}
%  \label{fig:initial_crack}
% \end{figure}
Our training data is comprised of two parts, one consisting of 450 snapshots from a crack propagation simulation with a fixed traction magnitude of $\sigma=4$ MPa, and the other consisting of randomized sinusoidal displacement fields and the corresponding $L_1$, $L_2$ fields computed by the PD operator. The sinusoidal subset contains 4,096 instances without fracture and is used to diversify the training data and mitigate overfitting on the crack data. %For validation, we use the eDAFNO models to compute $L_1$ and $L_2$. 
%Additional details of the data generation process are provided in Appendix~\ref{app:b3}. 
For testing, we use the trained models 
%in a dynamic solver as $\mathcal{L}(\bm{u})$ surrogates at each time step. Note that the topology evolution rule in testing remains the same physical damage model (see Appendix B.3), with the difference that it updates the topology using the displacement field obtained via the forces predicted by the trained eDAFNO model. We test the trained eDAFNO model 
in two scenarios with traction magnitudes different from what is used in training, which allows us to evaluate the generalizability of eDAFNO. Note that these tests are highly challenging for the trained models, as the predictions of the previous time step are used as the models' input in the current step, which leads to error accumulation as the simulation marches forward.

\begin{figure}[!h]\centering
\includegraphics[width=.6\linewidth]{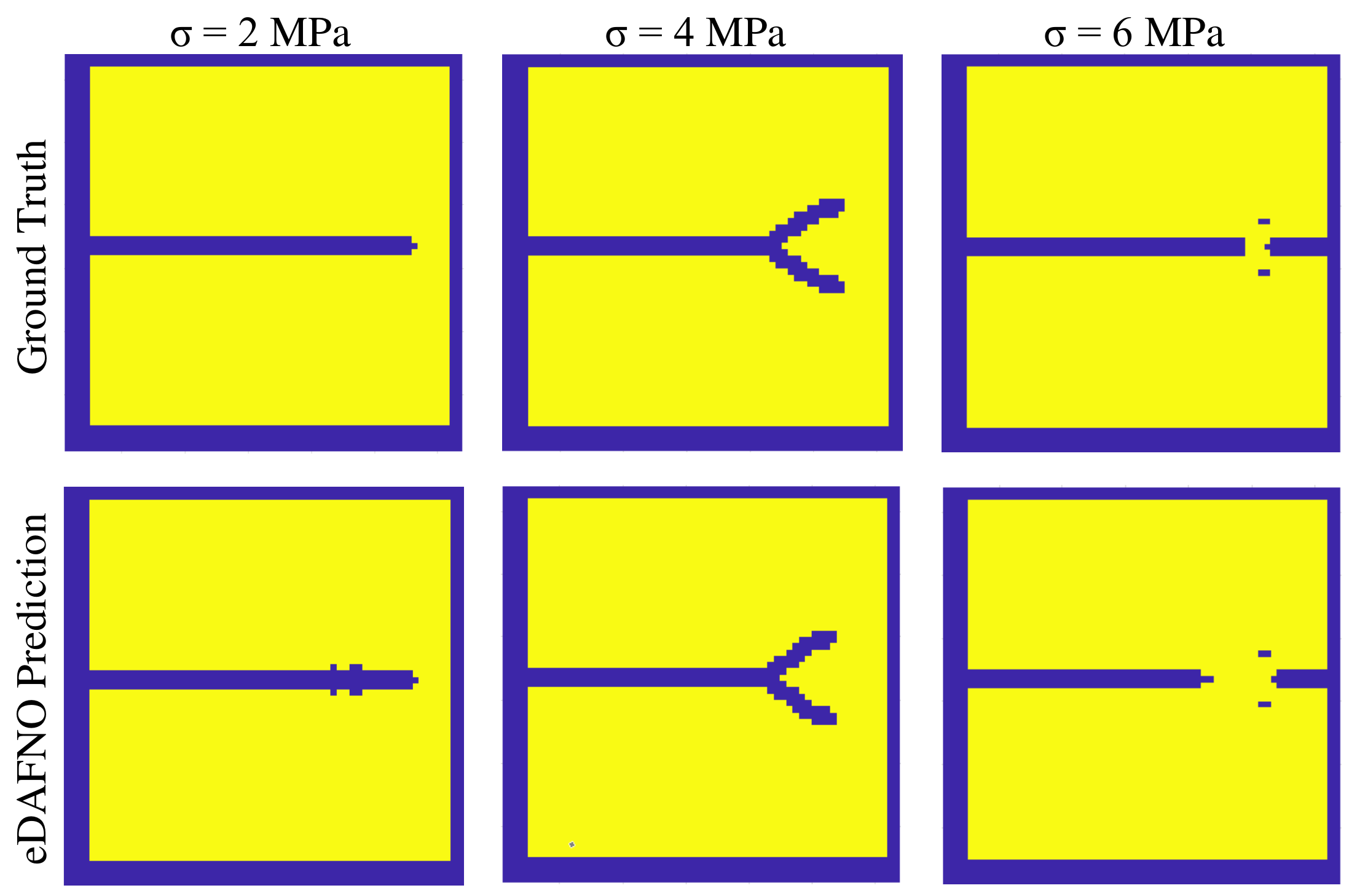}\vspace{-1mm}
 \caption{Comparison of the fracture patterns with different loading scenarios between the high-fidelity solution and eDAFNO prediction.\vspace{-1mm}}
 \label{fig:damage_prop}
\end{figure}

\begin{figure}[!h]\centering
\includegraphics[width=.49\linewidth]{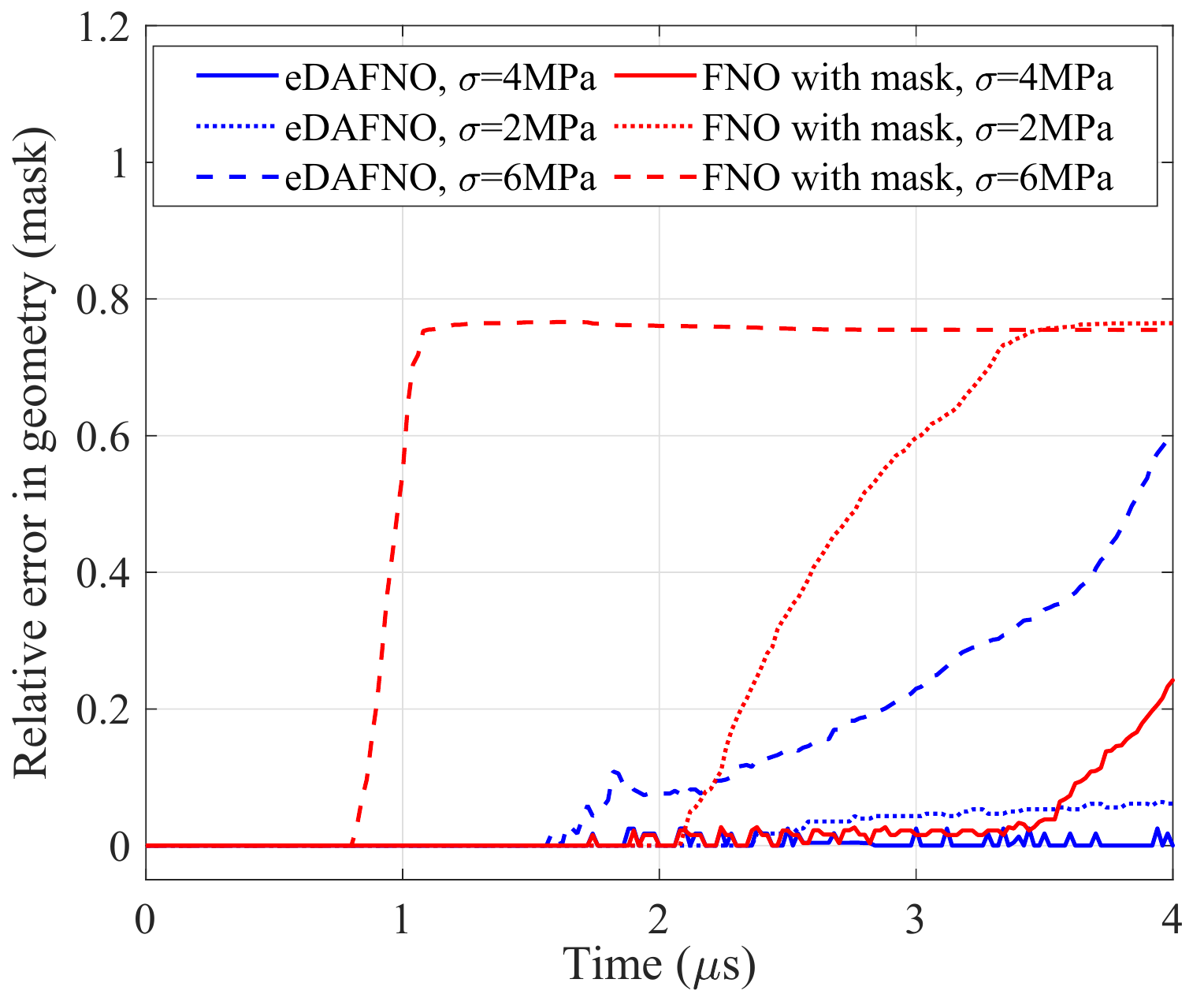}
\hspace{0.005\textwidth}
\includegraphics[width=.49\linewidth]{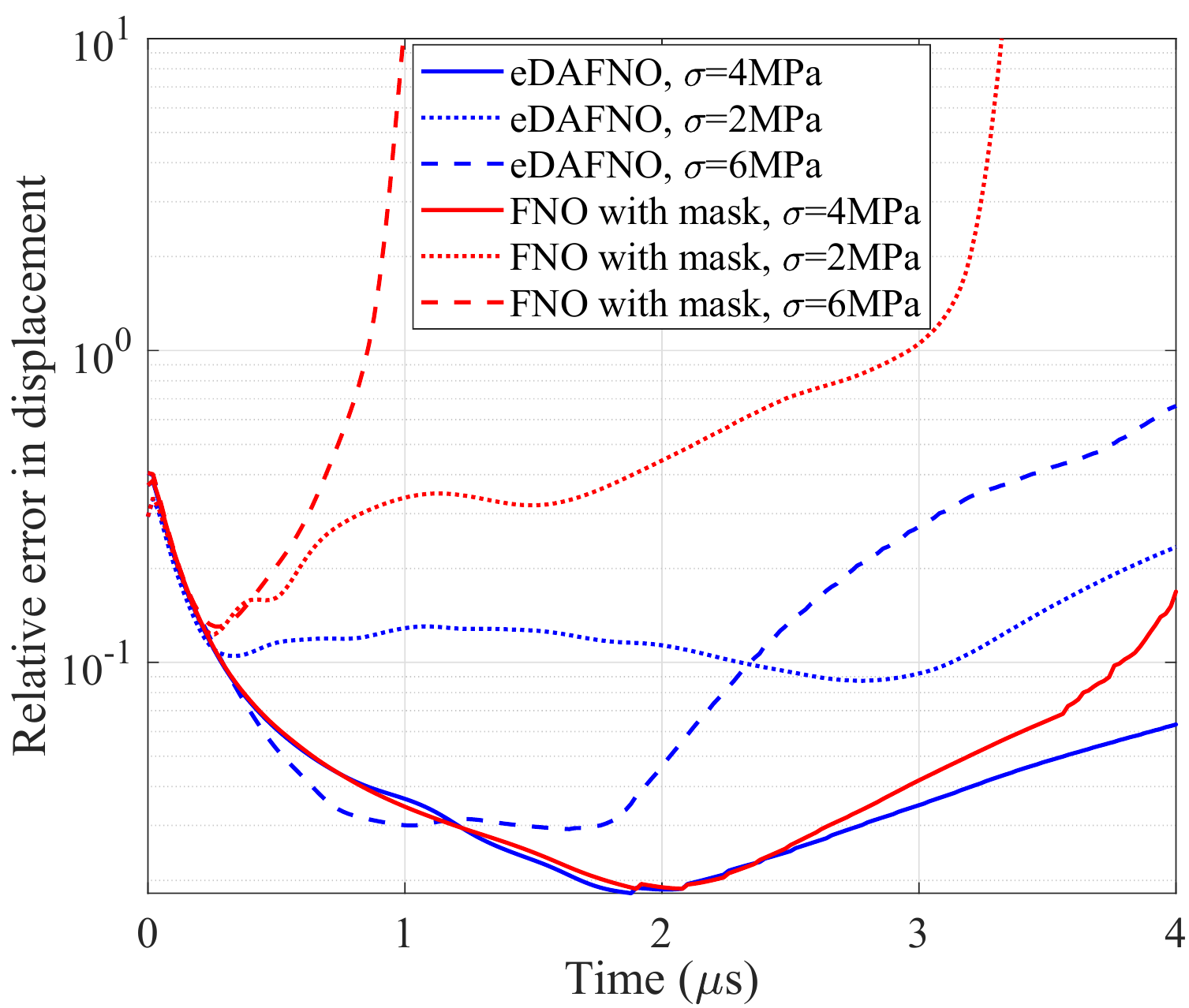}\vspace{-1mm}
\caption{Comparison of relative errors in the characteristic function $\chi$ (left) and displacement fields (right) of the evolving topology predicted using the trained eDAFNO and FNO with mask at each time step for the training (loading = 4 MPa) and testing (loading = 2 MPa and 6 MPa) scenarios.\vspace{-1mm}}
\label{fig:rel_errs}
\vspace{-3mm}
\end{figure}

% \begin{figure}[!t]\centering
% \includegraphics[width=.55\linewidth]{figures/error_damage.pdf}
%  \caption{Comparison of relative errors in the characteristic function $\chi$ of the evolving topology predicted using the trained eDAFNO and FNO with mask at each time step for the training (loading = 4 MPa) and testing (loading = 2 MPa and 6 MPa) scenarios.}
%  \label{fig:error-chi}
% \end{figure}

% \begin{figure}[!t]\centering
% \includegraphics[width=.55\linewidth]{figures/error_u.pdf}
%  \caption{Comparison of relative errors of the predicted displacement fields using the trained eDAFNO and FNO with mask at each time step for the training (loading = 4 MPa) and testing (loading = 2 MPa and 6 MPa) scenarios.}
%  \label{fig:error-u}
% \end{figure}

% \begin{figure}[!t]\centering
% \includegraphics[width=.65\linewidth]{figures/cross_resolution.pdf}
%  \caption{Demonstration of the resolution independence property of eDAFNO trained using a spatial discretization of 64$\times$64 and time step of 0.02$\mu$s and tested on a spatial discretization of 128$\times$128 and time step of 0.01$\mu$s. The three rows correspond to the $\chi$, $u_1$, and $u_2$ fields, respectively.}
%  \label{fig:cross_resolution}
% \end{figure}

Figure~\ref{fig:damage_prop} displays the test results on the crack patterns under different traction magnitudes, where the low traction magnitude (left column) shows a slow straight crack, the mid-level traction results in a crack with a moderate speed and a bifurcation event (middle column), and the highest traction leads to a rapid crack growth and initiates a second crack from the other side of the plate (right column). %as well as some distributed damage between the two cracks as they approach each other. 
The trained eDAFNO model is able to accurately predict the left and right examples while only seeing the middle one, indicating that it has correctly learned the true constitutive behavior of the material and generalized well to previously unseen loading scenarios and the correspondingly changed domain topology. To further verify eDAFNO's generalizability to unseen geometries, in Figure~\ref{fig:rel_errs} we compare eDAFNO with the baseline FNO w/ smoothed mask model, and plot their relative errors in $\chi$ and $\ub$. 
%Note that the damage in the data-driven model could potentially become unstable as the error accumulates. Nevertheless, the results in Figure~\ref{fig:damage_prop} are encouraging in the sense that the introduced modification in DAFNO is highly effective. Along with further improvements and better training strategies, DAFNO can become a great asset in constructing efficient and generalizable operators that are not restricted to a certain geometry.
%Figure~\ref{fig:rel_errs} reports the relative error in $\chi(t)$ and in the displacement field, respectively, at each time step for both the trained scenario (loading traction 4 MPa) and the test scenarios (loading traction = 2 MPa and 6 MPa). 
%We also test FNO with mask as an additional baseline, where the evolving geometry is updated via smoothed masks as an additional input to the FNO model. As observed, in each scenario eDAFNO outperforms FNO with mask. 
As observed, the FNO predictions become unstable at a fairly early time in the two test scenarios, since it creates a mask which is out of training distribution, while DAFNO can handle different crack patterns by hard coding it in the architecture, so the results remain stable for a much longer time.

% Another important advantage of DAFNO is its property of resolution-independence. In Figure~\ref{fig:cross_resolution}, we show the predictability of the trained eDAFNO across different resolutions. Specifically, the eDAFNO model is trained on a spatial resolution of 64×64 and a time step of 0.02 $\mu$s, and it is here tested on both a finer spatial resolution of 128×128 and a finer time step of 0.01 $\mu$s. The performance of the low-resolution-trained eDAFNO on high resolutions is compared with the high-fidelity peridynamics simulation results, where visually identical results are observed. Note that, although the time marching is computed with an ODE solver in this example, the temporal resolution independence is still worth investigating because, as the number of time steps increases, the number of times that the error accumulates in the dynamic solver increases as well. Our results show that eDAFNO prediction remains independent of the time step employed.

\section{Conclusion}
\vspace{-1mm}

We introduce two DAFNO variants to enable FNOs on irregular domains for PDE solution operator learning. By incorporating the geometric information from a smoothed characteristic function in the iterative Fourier layer while retaining the convolutional form, DAFNO possesses the computational efficiency from FFT together with the flexibility to operate on different computational domains. As a result, DAFNO is not only highly efficient and flexible in solving problems involving arbitrary shapes, but it also manifests its generalizability on evolving domains with topology change. In two benchmark datasets and a real-world crack propagation dataset, we demonstrate the state-of-the-art performance of DAFNO. 
We find both architectures helpful in practice: eDAFNO is slightly more accurate while iDAFNO is more computationally efficient and less overfitting with limited data. 

\textbf{Limitation:} Due to the requirement of the standard FFT package, in the current DAFNO we focus on changing domains with uniformly meshed grids. However, we point out that this limitation can be lifted by using nonuniform FFT \citep{greengard2004accelerating} or an additional mapping for grid deformation, as shown in the airfoil experiment. Additionally, in our applications, we have focused on applying the same types of boundary conditions to the changing domains (e.g., all Dirichlet or all Neumann). In this context, another limitation and possible future direction would be on the transferability to PDEs with different types of boundary conditions.

%We combine these two architectures with a tanh-smoothened boundary encoding technique to allow for state-of-the-art performance on tested examples.
% As future work, we plan to consider more applications involving complicated geometries.

% As future work, we plan to consider further applications involving complicated geometries, such as the object contact problems and the fluid--structure interaction problems.

% \section*{References}
\newpage

\begin{ack}
S. Jafarzadeh would like to acknowledge support by the AFOSR grant FA9550-22-1-0197, and Y. Yu would like to acknowledge support by the National Science Foundation under award DMS-1753031 and the AFOSR grant FA9550-22-1-0197. Portions of this research were conducted on Lehigh University's Research Computing infrastructure partially supported by NSF Award 2019035.
\end{ack}

\bibliography{nliu}
\bibliographystyle{icml2023}

\newpage
\appendix
\onecolumn
\section{Detailed iDAFNO architecture}\label{app:a}

Similar to the eDAFNO architecture shown in \eqref{eq:fno_with_chi}, we present the iDAFNO version by incorporating the layer-independent parameter definition characterized in the IFNO structure \citep{you2022learning}:
\begin{align}
\begin{split}
\mathcal{J}[\hb](\bm{x}) :=& \bm{h}(\bm{x}) + \tau \sigma \biggl(\chi(\bm{x}) \bigl(\mathcal{I}(\chi(\cdot) \bm{h}(\cdot);\vb)  -\bm{h}(\bm{x}) \mathcal{I}(\chi(\cdot);\vb) + W \bm{h}(\bm{x})+\cb \bigr)\biggr) \text{ ,}\\
\text{where }\;\; & \mathcal{I}(\circ;\vb) := \mathcal{F}^{-1} \bigl[\mathcal{F}[\kappa(\cdot;\vb)] \cdot \mathcal{F}[\circ]\bigr]\text{ .}
\label{eq:iDAFNO}
\raisetag{14pt}
\end{split}
\end{align}
Here, $\tau=\dfrac{1}{L}$ is the reciprocal of the total number of layers employed. Note that the superscript $l$ is dropped because the model parameters are layer-independent in the iDAFNO architecture, which leads to significant computational saving.

\section{Problem settings and additional experimental results}\label{app:b}

%\begin{figure}[!t]\centering
%\includegraphics[width=.5\linewidth]{figures/topology_change.png}
% \caption{An illustration of crack propagation modeling as a topology change in DAFNO.}
% \label{fig:topology_change}
%\end{figure}

In order to maintain consistency with other baselines, the dimension of representation in the first two examples is set to $d_h=32$, with a total of 4 Fourier layers and 12 Fourier modes being used in each direction. The output at each point is obtained via a projection layer in the form of a 2-layer multilayer perceptron (MLP) with width $(d_h, 128, d_u)$, where $d_u$ is the intended number of output. In the third example, $d_h$ is set to 16. A total of 3 Fourier layers with 32 Fourier modes in each direction are employed. The width of the projection MLP is set to $(d_h, 2d_h, d_u)$.

\subsection{Experiment 1 -- Constitutive modeling of hyperelasticity}\label{app:b1}

The dataset is obtained from \citet{li2022fourier}, which consists of an interpolated dataset of $41\times41$ point cloud on uniformly structured grids. 
%An example of the computational domain, the smoothened boundary encoding, the ground truth, and the eDAFNO prediction is illustrated in Figure~\ref{fig:elas}. 
The parameter of each method is given in the following, where the parameter choice of each model is selected by tuning the number of layers and the width (channel dimension) keeping the total number of parameters on the same magnitude. 
\begin{itemize}
\item eDAFNO: In these cases, we use neural operators to construct mapping from grid location $\xb$ as the input, and the stress field as the output. To perform fair comparison with the results reported in \cite{li2022fourier}, we employ the same hyperparameters here: in particular, four Fourier layers with mode
12 and width 32 are used.
\item iDAFNO: The iDAFNO cases employ the same hyperparameters as the eDAFNO cases, with the iterative layer structure demonstrated in \eqref{eq:iDAFNO}. In iDAFNO, all Fourier layers share the same set of trainable parameters, while different layers have different parameters in eDAFNO. Hence, iDAFNO reduces the number of trainable parameters by almost $75\%$, when using the same hyperparameters as in eDAFNO.
\item FNO (with mask or smooth $\chi$): Following the same practice as in \cite{li2022fourier}, we train a plain FNO model \citep{li2020fourier}, with the input as $[\xb,\chi(\xb)]$ (in the ``with mask'' cases) or as $[\xb,\tilde{\chi}(\xb)]$ (in the ``with smoothed $\chi$'' cases). Herein, we employ the same FNO architecture as reported in \cite{li2022fourier}, where four Fourier layers are used with mode 12 and width 32.
\item IFNO (with mask or smooth $\chi$): Similar to the FNO cases, we also use $[\xb,\chi(\xb)]$ (in the ``with mask'' cases) or $[\xb,\tilde{\chi}(\xb)]$ (in the ``with smoothed $\chi$'' cases) as the input, with four Fourier layers, mode 12, and width 32. On the Fourier layers, the implicit architecture proposed in \cite{you2022learning} is employed, such that all four Fourier layers share the same set of trainable parameters. Therefore, the number of trainable parameters in IFNOs is roughly 1/4 of that in FNOs.
\item Geo-FNO: As a baseline model for FNOs with various geometries, we employ the Geo-FNO architecture from \cite{li2022fourier}, where an additional deformation neural network is trained together with FNO to provide a diffeomorphism from uniform grids to the deformed domain.
\item F-FNO: Following the settings in \cite{li2022fourier}, we train the F-FNO model \citep{li2020fourier} with the input $[\xb,\chi(\xb)]$. We adopt the same F-FNO architecture as reported in \cite{tran2022factorized}, where four Fourier layers are used with mode 16 and width 64.
\item GNO: The graph neural operators are flexible on the problem geometry, which have been widely used for complex geometries \citep{li2020neural,liu2022ino}. To carry out fair comparison, we build a full graph with edge connection radius $r$ = 0.2, width 32 and kernel width 512. As a result, the total number of parameters in GNOs is on the same magnitude as in FNOs.
\item DeepONet: As another neural operator baseline model, the deep operator network \citep{lu2022comprehensive} is composed of two neural networks -- a trunk net and a branch net to represent the basis and coefficients of the operator. In this baseline, we use five layers for both the trunk net and branch net, each with a width of 256.
\item UNet: Analogous to the setup in \cite{li2022fourier}, we train a UNet model \citep{ronneberger2015u} on uniform grids, where 4 downsampling and upsampling blocks with 20 hidden channels are employed.
\end{itemize}

The comparison of the total number of parameters of the selected models used in the hyperelasticity problem is listed in Table~\ref{tab:elas_para}\footnote{We note that the numbers of trainable parameters for the ``Geo-FNO'' and ``FNO'' cases are different from the ones provided in \cite{li2022fourier}. For fair comparison with methods using real-valued trainable parameters, we count each complex-valued trainable parameter as two degrees of freedom.}. In addition, the average runtime for each method on the hyperelasticity problem with 1000 training samples is provided in Table~\ref{tab:elas_runtime}. All tests are performed on a NVIDIA RTX A6000 GPU card with 48GB memory. From this table, we can see that in DAFNOs the runtime increases slightly compared with the corresponding FNOs, but they are still substantially more efficient than other baselines, such as Geo-FNO.

\begin{figure}[!h]\centering
\includegraphics[width=.6\linewidth]{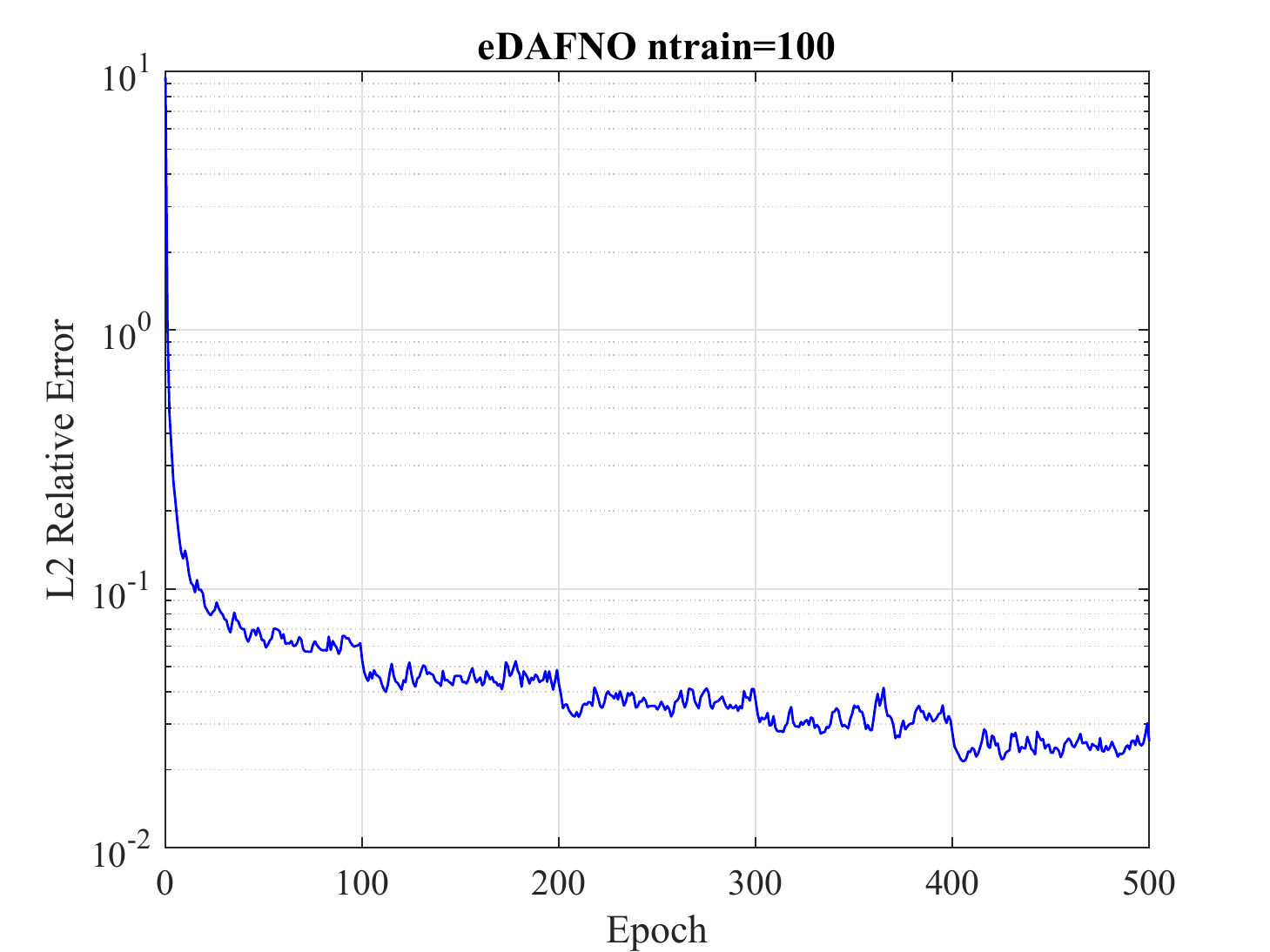}
 \caption{Demonstration of a typical training curve for eDAFNO.}
 \label{fig:train_curve}
\end{figure}

For each method, we tune the learning rate from the range [1e-3,1e-1], the decay rate from the range [0.4,0.9], the weight decay parameter from from the range [1e-6,1e-2], and the smoothing coefficient (where applicable) from the range [5,100], then report the model with the best validation error.  A typical training curve can be found in Figure~\ref{fig:train_curve}. As a supplement of Table~\ref{tab:elas_results}, the full table of all training and testing errors from different models is provided in Table~\ref{tab:elas_resultsmore}.

\begin{table}[!h]
    \caption{The per-epoch runtime (in seconds) of selected models for the hyperelasticity problem.}
    \label{tab:elas_runtime}
    \centering
    {\small 
    \begin{tabular}{cccccccccc}
    \toprule
         model & eDAFNO & iDAFNO & FNO & IFNO & Geo-FNO & GNO & DeepONet & UNet & F-FNO\\
         \midrule
         runtime & 2.00 & 1.70 & 1.81 & 1.62 & 5.12 & 98.37 & 940.12 & 5.04 & 3.41 \\
         \bottomrule
\end{tabular}}
\end{table}

\begin{table}[h!]
\caption{Results for the hyperelasticity problem, where bold numbers highlight the best method according to the test error.}% The experiments of each method are repeated on 5 trials with 5 different random seeds.}
\label{tab:elas_resultsmore}
\begin{center}
{\small    \centering
%\begin{tabular}{|c|c|c|c|}
\begin{tabular}{l|lll}
    \toprule
        {Model, Dataset} & \multicolumn{3}{c}{\# of training samples} \\
        \cline{2-4}
        & 10 & 100 & 1000 \\
        \hline \hline
eDAFNO, train & 6.800\%$\pm$0.670\% & 2.050\%$\pm$0.035\% & 0.664\%$\pm$0.014\%\\
eDAFNO, test & \bf{16.446}\%$\pm$\bf{0.472}\% & 4.247\%$\pm$0.066\% & \bf{1.094}\%$\pm$\bf{0.012}\%\\
\hline
iDAFNO, train & 7.266\%$\pm$0.923\% & 2.038\%$\pm$0.036\% & 0.812\%$\pm$0.012\%\\
iDAFNO, test & 16.669\%$\pm$0.523\% & \bf{4.214}\%$\pm$\bf{0.058}\% & 1.207\%$\pm$0.006\%\\
\hline
FNO w/ mask, train & 
         2.907\%$\pm$0.318\%&2.277\%$\pm$0.240\% &0.881\%$\pm$0.015\%\\
FNO w/ mask, test & 
         19.487\%$\pm$0.633\%& 7.852\%$\pm$0.130\% & 4.550\%$\pm$0.062\%\\
\hline
FNO w/ smooth $\chi$, train & 
          2.876\%$\pm$0.152\% & 2.058\%$\pm$0.132\%& 0.815\%$\pm$0.012\%\\
FNO w/ smooth $\chi$, test & 
         17.431\%$\pm$0.536\%&5.479\%$\pm$0.186\% & 1.415\%$\pm$0.025\%\\
\hline
Geo-FNO, train & 
        0.547\%$\pm$0.336\%&0.689\%$\pm$0.676\% &1.192\%$\pm$0.232\%\\
         Geo-FNO, test & 
         28.725\%$\pm$2.600\%&10.343\%$\pm$4.446\% &2.316\%$\pm$0.283\%\\
\hline
IFNO w/ mask, train & 2.274\%$\pm$0.248\% & 1.687\%$\pm$0.047\% & 2.701\%$\pm$0.041\%\\
IFNO w/ mask, test & 19.262\%$\pm$0.376\%& 7.700\%$\pm$0.062\%& 4.481\%$\pm$0.022\%\\
\hline
IFNO w/ smooth $\chi$, train & 3.704\%$\pm$0.299\% & 1.683\%$\pm$0.029\% & 1.013\%$\pm$0.014\%\\
IFNO w/ smooth $\chi$, test & 17.145\%$\pm$0.432\% & 5.088\%$\pm$0.146\%& 1.509\%$\pm$0.018\%\\
\hline
GNO, train & 27.337\%$\pm$0.501\% & 18.713\%$\pm$0.669\% & 13.321\%$\pm$0.681\%\\
GNO, test & 29.305\%$\pm$0.321\% & 18.574\%$\pm$0.584\% & 13.007\%$\pm$0.729\%\\
\hline
DeepONet, train & 23.071\%$\pm$5.963\% & 22.700\%$\pm$0.984\% & 7.937\%$\pm$0.309\%\\
DeepONet, test & 35.409\%$\pm$0.408\% & 25.925\%$\pm$0.724\% & 11.760\%$\pm$0.827\%\\
\hline
UNet, train & 98.042\%$\pm$0.260\% & 34.569\%$\pm$2.676\% & 1.760\%$\pm$0.115\%\\
UNet, test & 98.167\%$\pm$0.236\% & 34.467\%$\pm$2.858\% & 5.462\%$\pm$0.048\%\\
\bottomrule
\end{tabular}}
\end{center}
\end{table}

% \begin{figure}[!h]\centering
% \includegraphics[width=.99\linewidth]{figures/elas_demo.pdf}
%  \caption{Constitutive modeling of hyperelasticity: an illustration of the discretized computational domain, the smoothened boundary encoding (i.e., smooth $\chi$), the ground truth, and the eDAFNO prediction.}
%  \label{fig:elas}
% \end{figure}

%To further test the effect of the proposed smoothed characteristic function on the resolution independence of the model (particularly at higher resolutions), a zero-shot super-resolution test is performed using an eDAFNO model trained on $41\times41$ grids and directly evaluated on a $161\times161$ grid. The results are displayed in Figure~\ref{fig:super_resolution}, where no notable artifact is observed.

% \begin{figure}[!h]\centering
% \includegraphics[width=.99\linewidth]{figures/elas_super_resolution.png}
%  \caption{Zero-shot super resolution: the eDAFNO model is trained using 41$\times$41 discretization and evaluated directly on 161$\times$161 discretization.}
%  \label{fig:super_resolution}
% \end{figure}

\begin{figure}[!t]\centering
\includegraphics[width=1.0\linewidth]{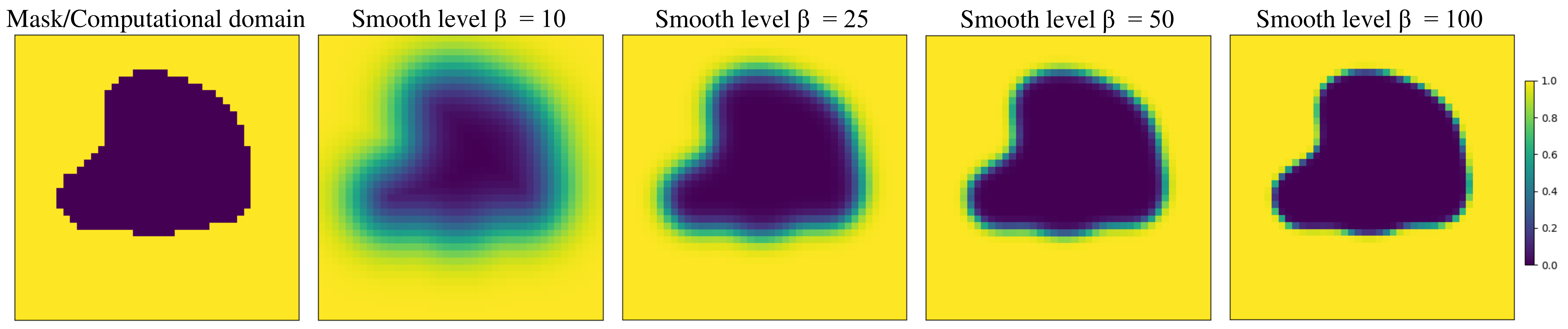}
 \caption{An illustration of the effect of varying the smoothing coefficient on the resulting boundary encoding. The larger the smoothing level $\beta$ is, the sharper and narrower the encoded boundary becomes. In effect, $\beta$ can be treated as a hyperparameter and tuned according to the validation error to either smoothen the boundary or keep the original boundary untouched.}
 \label{fig:boundary_encoding}
\end{figure}

To demonstrate the effect of the smoothing level when using different smoothing coefficient $\beta$, we illustrate the smoothed $\chi$ on an exemplar test sample in Figure \ref{fig:boundary_encoding}. We also perform tests on the hyperelasticity example with a total of 1000 training samples and show the errors on the test dataset in Table~\ref{tab:elas_beta_ablation}. For each value of $\beta$, we search for the optimal initial learning rate, the decay rate, and the weight decay parameter based on the validation dataset, and report the optimal values.

\begin{table}[h!]
    \caption{The effect of the smoothing coefficient $\beta$ on test loss in the hyperelasticity example with a total of 1000 training samples.}
    \label{tab:elas_beta_ablation}
    \centering
    {\small \centering
    \begin{tabular}{cccccc}
    \toprule
         initial learning rate & decay rate & weight decay parameter & $\beta$ & train loss & test loss\\
         \midrule
         $4.5\times10^{-2}$ & 0.5 & $3\times10^{-6}$ & 5 & 0.564\% & 1.155\%\\
         %\hline 
         $4.0\times10^{-2}$ & 0.5 & $1\times10^{-5}$ & 10 & 0.637\% & 1.064\%\\
         %\hline 
         $2.0\times10^{-2}$ & 0.5 & $1\times10^{-5}$ & 20 & 0.454\% & 1.120\%\\
         %\hline 
         $1.5\times10^{-2}$ & 0.5 & $3\times10^{-5}$ & 30 & 0.516\% & 1.147\%\\
         %\hline 
         $2.5\times10^{-2}$ & 0.5 & $3\times10^{-5}$ & 40 & 0.608\% & 1.135\%\\
         %\hline 
         $1.5\times10^{-2}$ & 0.5 & $3\times10^{-5}$ & 50 & 0.504\% & 1.179\%\\
         %\hline 
         $1.5\times10^{-2}$ & 0.5 & $2\times10^{-5}$ & 60 & 0.498\% & 1.194\%\\
         %\hline 
         $1.5\times10^{-2}$ & 0.5 & $3\times10^{-5}$ & 70 & 0.508\% & 1.240\%\\
         %\hline 
         $1.5\times10^{-2}$ & 0.5 & $3\times10^{-5}$ & 80 & 0.515\% & 1.275\%\\
         %\hline 
         $1.5\times10^{-2}$ & 0.5 & $3\times10^{-5}$ & 90 & 0.529\% & 1.306\%\\
         %\hline 
         $3.0\times10^{-2}$ & 0.5 & $3\times10^{-5}$ & 100 & 0.675\% & 1.338\%\\
         \bottomrule
    \end{tabular}}
\end{table}

\subsection{Experiment 2 -- Airfoil design}\label{app:b2}
The airfoil dataset is directly taken from \citet{li2022fourier}, which is an interpolated dataset of $101\times101$ point cloud on uniformly structured grids. The analytical mapping function $f$ and the corresponding inverse mapping function $f^{-1}$ used in Figure~\ref{fig:airfoil_grids_deformation} are defined in the following:
\begin{align}
\begin{bmatrix}
x\\
y
\end{bmatrix}=&f\left(\begin{bmatrix}
X\\
Y
\end{bmatrix}\right) = \begin{bmatrix}
0.909 \tan^{-1}\left(1.965X\right)\\
0.714\tan^{-1}\left(3.46Y+0.173\sin\left(0.909\pi \tan^{-1}\left(1.965X\right)\right)\right)
\end{bmatrix}\text{ ,}\\
\begin{bmatrix}
X\\
Y
\end{bmatrix}=&f^{-1}\left(\begin{bmatrix}
x\\
y
\end{bmatrix}\right) = \begin{bmatrix}
0.509 \tan\left(1.1x\right)\\
0.289\tan\left(1.4y\right)-0.05\sin\left(\pi x\right)
\end{bmatrix}
\text{ ,}
\end{align}
where upper- and lower-case letters indicate the coordinate systems in the physical and computational spaces, respectively.

\subsection{Experiment 3 -- Crack propagation with topology change}\label{app:b3}

\begin{figure}[h!]\centering
\includegraphics[width=.6\linewidth]{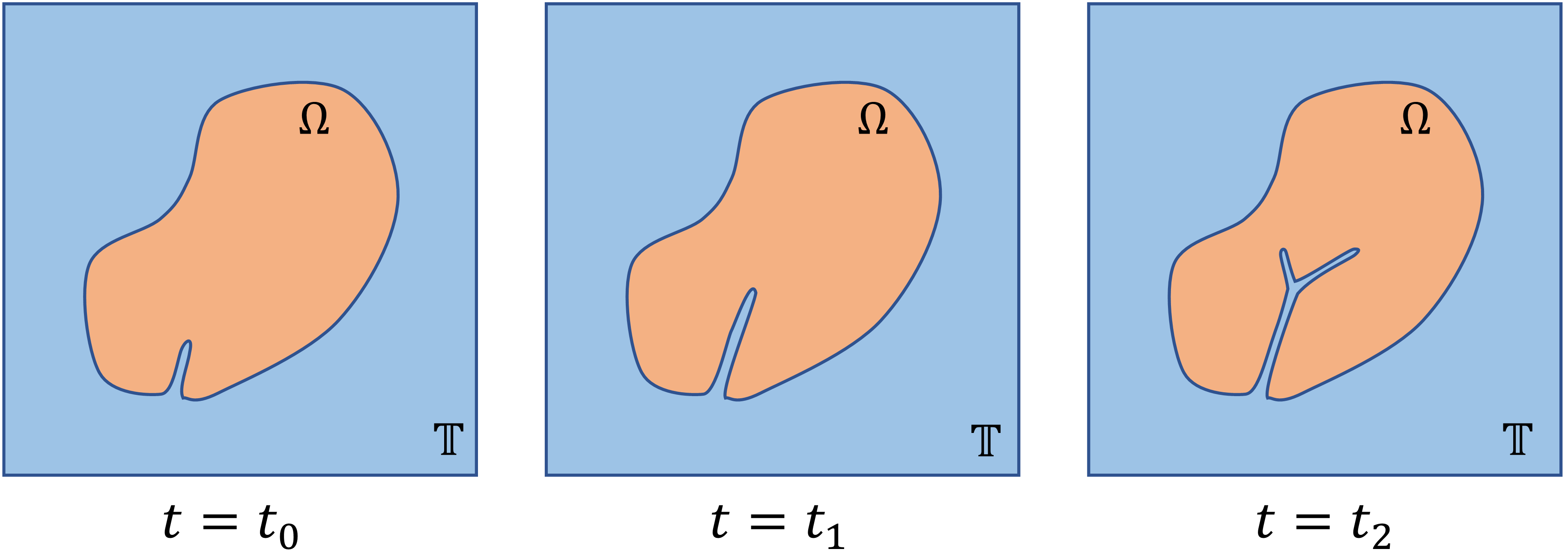}\qquad
\includegraphics[width=.18\linewidth]{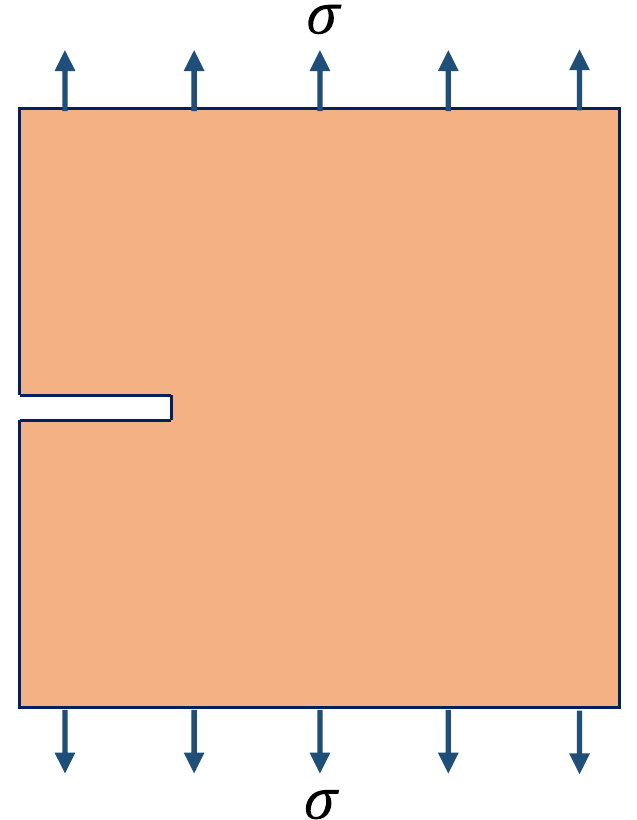}
 \caption{An illustration of the crack propagation problem, showing the topology change in DAFNO (left) and the physical problem setup (right), where a 2D plate with a pre-crack is subjected to external tractions (denoted as $\sigma$ with a slight abuse of notation) on the top and bottom edges.}
 \label{fig:topology_change}
\end{figure}

An illustration of the time-dependent domain evolution, as well as the problem setup, is shown in Figure~\ref{fig:topology_change}. The governing PD equation of motion for brittle fracture used in generating the dataset is given below:
\begin{equation}\label{eq:pd}
\rho \frac{\partial^2 \bm{u}}{\partial t^2} = \mathcal{L}(\bm{u})+\bm{b} \text{ ,  } \mathcal{L}(\bm{u})(\bm{x}, t) = \int_{\mathcal{H}_x}\mu(\bm{x}, \bm{y}, t) f(\bm{x}, \bm{y}, t)d\bm{y}\text{ .}
\end{equation}
In \eqref{eq:pd}, $\bm{u}$ is the displacement field, $\rho$ is mass density, $t$ is time, $\bm{b}$ is the external force density (a.k.a. the body force), and $\mathcal{L}(\bm{u})$ is the internal force density. $\mathcal{L}(\bm{u})$ is the divergence of stress in local theory, but in PD, it is defined by the integral described in \eqref{eq:pd}. $\mathcal{H}_x$ denotes a finite-size neighborhood of point $\bm{x}$. $f(\bm{x}, \bm{y}, t)$ is the dual force density representing the pairwise force acting between unit volumes at points $\bm{x}$ and $\bm{y}$ in its neighborhood $\mathcal{H}_x$. $f$ depends on the PD constitutive model, and $\mu(\bm{x}, \bm{y}, t)$ is a binary history-dependent quantity representing material damage. $\mu$ is either 0 or 1 for brittle fracture models, where $\mu=0$ denotes a lost interaction for material points $\bm{x}$ and $\bm{y}$ while $\mu=1$ implies an intact connection between the two. For the material model, we choose to work with the linearized bond-based model, and for the damage model we adopt the pointwise energy-based model provided in the PeriFast software. According to the employed damage model, topology evolution due to growing crack is a function of strain energy which depends on the updated displacement field, i.e., $\Omega(t) = \Omega(\bm{u}(t))$. Additional details regarding the PD formulation can be found in \citet{jafarzadeh2022general}.

The physical parameters used in generating the data are: Young's modulus $E=150$ GPa, Poison's ratio $\nu=0.33$, mass density $\rho=1000 \text{ kg/m}^3$, and fracture energy $G_0=200 \text{ J/m}^2$. The relative computational parameters are: PD horizon (the radius of the neighborhood for nonlocal interactions) $\delta=2.07$ mm, extended domain (i.e., the periodic box) size $44.14\text{ mm}\times44.14 \text{ mm}$ with $64\times64$ discretization, and time step $\Delta t=2\times10^{-8}$ s. For the crack data subset in training, we run PeriFast software with the above parameters and the traction magnitude of 4 MPa. We record $u_1$, $u_2$, $\chi$, $L_1$, and $L_2$ for 450 consecutive time steps. For the sinusoidal data subset used in training, we set $u_1=c \sin\left(\frac{2m\pi x_1}{L}\right)\sin\left(\frac{2n\pi x_2}{L}\right)$, and $u_2=0$ for $m,n=1,2,\cdots,32$, where $L$ is the length of the square box, $x_1$ and $x_2$ are the 2D coordinates, and $c=0.01/32$ is a scaling factor to make the generated displacement in the same scale as the crack data. Additionally, we set $u_1=0$ and $u_2=c \sin\left(\frac{2m\pi x_1}{L}\right)\sin\left(\frac{2n\pi x_2}{L}\right)$. This results in a total of 2,048 instances of sinusoidal displacement fields. Next, we set $\chi=1$ for all nodes and use the PD operator in PeriFast to compute the corresponding $L_1$ and $L_2$ fields. 
%The sinusoidal wave displacement fields in various modes without crack help the DAFNO models learn the constitutive behavior and mitigate overfitting on the crack data.

Following the common practice in PD simulations \citep{ha2010studies}, we employ the following two additional techniques to help with training and stabilizing crack propagation. Firstly, we do not allow damage to initiate from boundaries. This technique has been used in previous PD simulations and is referred to as the ``no-fail zone''. It effectively stops unrealistic distributed damage from initiating on the boundaries. Secondly, given that the physical problem is symmetric, we enforce the damage growth in the simulation with eDAFNO to be symmetric as well. Note that the whole domain is used for training, and the predictions on the entire domain is used for next time step evaluation. Symmetry is enforced only when the topology characteristic function $\chi$ is updated.

%\section{Additional experimental results}
\begin{figure}[h!]\centering
\includegraphics[width=.7\linewidth]{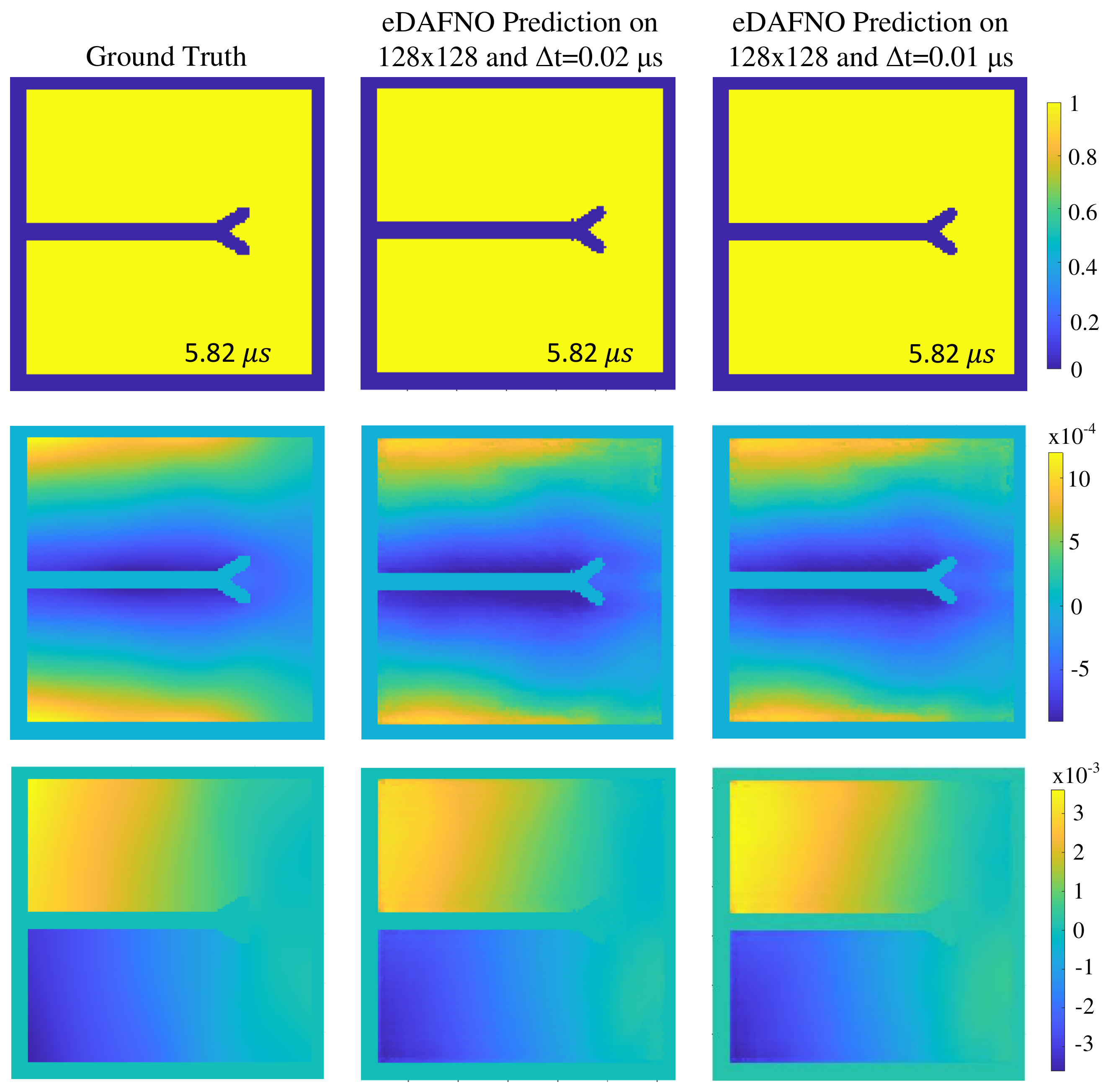}
 \caption{Demonstration of the resolution-independence property of eDAFNO trained using a spatial discretization of 64$\times$64 and time step of 0.02 $\mu$s and tested on a spatial discretization of 128$\times$128 and time step of 0.01 $\mu$s. The three rows correspond to the $\chi$, $u_1$, and $u_2$ fields, respectively.}
 \label{fig:cross_resolution}
\end{figure}

Besides the resolution-independence property of DAFNO as shown in Figure~\ref{fig:super_resolution}, we further investigate the generalizability of DAFNO in both physical and temporal resolutions with this example. %In Figure~\ref{fig:cross_resolution}, we show the predictability of the trained eDAFNO across different resolutions. 
Specifically, the eDAFNO model is trained on a spatial resolution of 64×64 and a time step of 0.02 $\mu$s, and it is here tested on both a finer spatial resolution of 128×128 and a finer time step of 0.01 $\mu$s. As shown in Figure~\ref{fig:cross_resolution}, the performance of the low-resolution-trained eDAFNO on high resolutions is compared with the high-fidelity peridynamics simulation results, where visually identical results are observed. Note that, although the time marching is computed with an ODE solver in this example, the temporal resolution independence is still worth investigating because, as the number of time steps increases, the number of times that the error accumulates in the dynamic solver increases as well. Our results show that eDAFNO prediction remains independent of the time step employed.

\subsection{Experiment 4 -- Pipe flow}\label{app:b4}

We perform an additional experiment of pipe flow, in which the dataset is obtained from \cite{li2022fourier}. We closely follow their problem setup and briefly document the comparison against Geo-FNO in what follows. 

\begin{figure}[!h]\centering
\includegraphics[width=.5\linewidth]{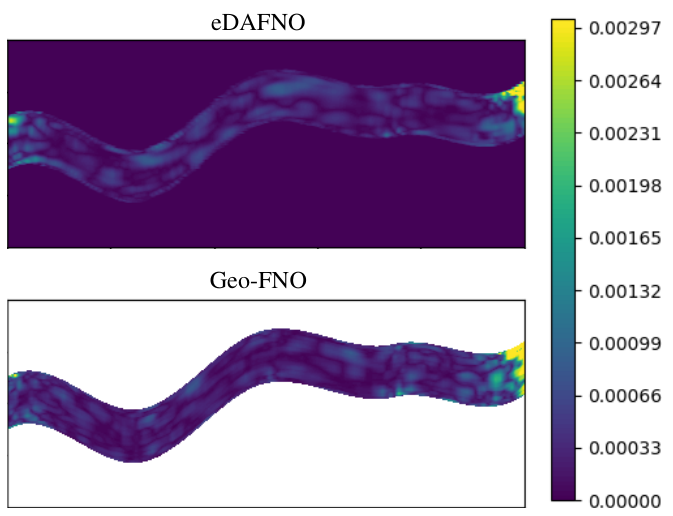}\vspace{1mm}
 \caption{An illustration of the absolute error distribution of predictions from Geo-FNO and eDAFNO on the pipe dataset. The maximum absolute errors of Geo-FNO and eDAFNO are 0.061 and 0.051, respectively. In the vicinity of the outlet where most errors accumulate, eDAFNO is also more accurate compared to Geo-FNO.}
 \label{fig:abs_error_pipe}
\end{figure}

Using 1000 samples for training, eDAFNO has achieved a similar performance to Geo-FNO: when comparing the relative $L^2$ errors, eDAFNO's test error on the pipe dataset is 0.71\%, while the test error of Geo-FNO is 0.67\%. When comparing the maximum absolute error (cf. Figure~\ref{fig:abs_error_pipe}), eDAFNO has $0.051$, while Geo-FNO has $0.061$. This is probably due to the fact that all pipes have a very simple geometry, which can be accurately represented with the pre-specified mapping in Geo-FNO. Note that such a pre-specified mapping for grid deformation can be easily added in DAFNO, as demonstrated in the airfoil experiment. In this circumstance, DAFNO becomes exactly the same as Geo-FNO in the pipe flow setup.

%%%%%%%%%%%%%%%%%%%%%%%%%%%%%%%%%%%%%%%%%%%%%%%%%%%%%%%%%%%%

\end{document}